\documentclass[letterpaper]{article} 
\usepackage{aaai2027}  
\usepackage[hyphens]{url}  
\usepackage{graphicx} 
\usepackage{natbib}  
\usepackage{caption} 
\usepackage{amsmath}
\usepackage{amssymb}
\usepackage{algorithm}
\usepackage{algorithmic}

\usepackage{newfloat}
\usepackage{listings}
\DeclareCaptionStyle{ruled}{labelfont=normalfont,labelsep=colon,strut=off} 
\floatstyle{ruled}
\newfloat{listing}{tb}{lst}{}
\floatname{listing}{Listing}

\usepackage{booktabs}

\usepackage{multirow}
\nocopyright 

\title{MoHallBench: A Benchmark for Motion Hallucination in Video Large Language Models\thanks{Preprint. Under review.}}

\author{
    Sihan Chen\textsuperscript{\rm 1,\rm 2}\equalcontrib,
    Jiale Li\textsuperscript{\rm 1}\equalcontrib,
    Jianghang Lin\textsuperscript{\rm 3},
    Mengyuan Liu\textsuperscript{\rm 1}\corresponding
}

\affiliations{
    \textsuperscript{\rm 1}
    State Key Laboratory of General Artificial Intelligence,
    Peking University, Shenzhen Graduate School,
    Shenzhen, China\\

    \textsuperscript{\rm 2}
    South China University of Technology,
    Guangzhou, China\\

    \textsuperscript{\rm 3}
    Key Laboratory of Multimedia Trusted Perception and Efficient Computing,
    Ministry of Education of China, Xiamen University,
    Xiamen, China\\

}

\begin{document}

\maketitle

\begin{abstract}
Video Large Language Models (VideoLLMs) have shown strong progress in video understanding, yet they still suffer from hallucinations that are inconsistent with visual evidence. 
Existing benchmarks mainly focus on object hallucination or coarse action perception, leaving a key video-specific problem underexplored: motion hallucination, in which models infer human motions that are absent from the video.
We present \textbf{MoHallBench}, a benchmark for diagnosing motion hallucination in VideoLLMs. 
MoHallBench systematically evaluates three major sources of hallucination: co-occurrence priors, sequential inference, and similarity confusion. 
It contains 11,306 video clips and 40,493 question-answer pairs, covering binary-choice, multiple-choice, and generative settings. 
We further introduce a bi-directional questioning protocol with bias-aware metrics to reduce affirmation bias in binary evaluation. 
Experiments on ten recent open-source VideoLLMs reveal a clear decoupling between action recognition and hallucination resistance, as models that perform well on positive action recognition often fail on adversarial negatives. 
Among all settings, sequential inference hallucination is the most severe, showing that current models tend to over-infer expected outcomes from partial motion cues. 
Our analyses further confirm that stronger priors and finer-grained similarity substantially amplify hallucination. We hope MoHallBench can facilitate future evaluation and mitigation of motion hallucination in VideoLLMs.
\end{abstract}

\section{Introduction}
\begin{figure*}[!t]
  \centering
  \includegraphics[width=\textwidth]{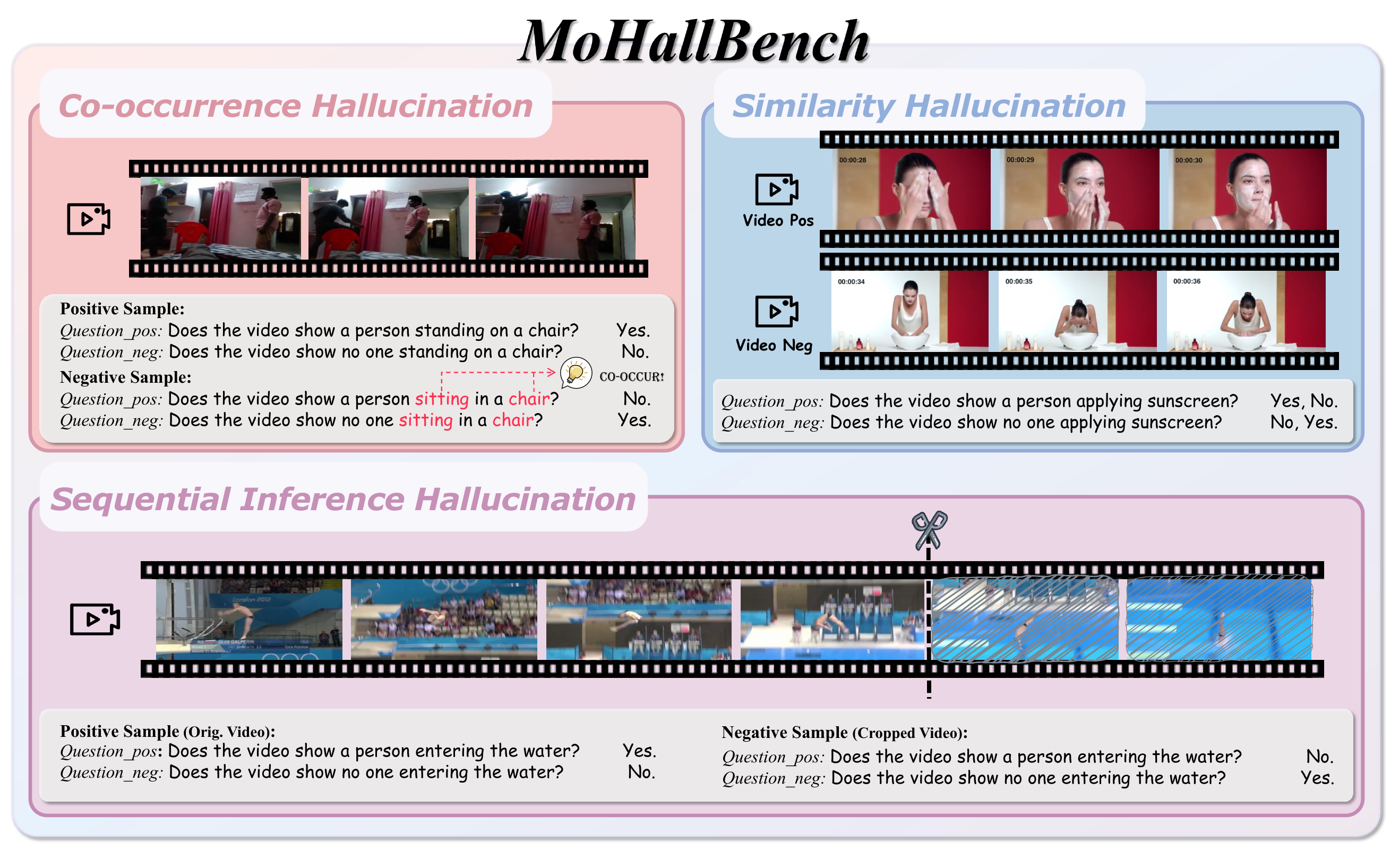}
  \caption{Overview of MoHallBench. We evaluate motion hallucination in VideoLLMs along three complementary axes: co-occurrence hallucination (CH), sequential inference hallucination (SIH), and similarity hallucination (SH).}
  \label{fig:figure1}
\end{figure*}
Recent progress in multimodal large language models (MLLMs) has rapidly expanded from image understanding~\cite{r26llava, r29instructblip, r28blip2, r40valley} to video understanding~\cite{r9qwen3vl, r15lnq, r12molmo2, r17internvl3, r18minicpm, r13vlm3, r14tarsier2, r16plm}. 
However, this progress is accompanied by the persistent challenge of multimodal hallucination where models generate plausible descriptions that are inconsistent with the visual input~\cite{r24hallusurvey, r25hallusurvey2}. 
Although hallucination in image settings has been extensively studied, the issue is becoming more pronounced in video models~\cite{r42motionbench, r33mvbench}. 
Compared with static images, videos contain an additional temporal dimension that introduces richer and more complex dynamics. 
Consequently, existing evaluations of object hallucination~\cite{r19pope, r31amber, r41chair}, namely whether a model describes nonexistent objects, are insufficient to capture the unique temporal characteristics of video data, leaving a gap in our understanding of how these models perceive and reason about dynamic human actions.
Moreover, existing video hallucination benchmarks capture only partial failure modes and do not systematically disentangle the causes~\cite{r21eventhallusioin, r22videohallucier, r23videhalluc, r20mhbench, r8mash}.
We therefore shift our attention to motion hallucination: whether a VideoLLM hallucinates human motions that are absent from the video. 

Understanding human motion requires deep spatiotemporal perception and fine-grained motion discrimination that go beyond surface-level recognition.
Spatially, an action is often defined by its interaction with specific objects~\cite{r30actionobject} and its occurrence within a relevant scene context~\cite{r8mash}. 
Temporally, actions frequently exhibit transition regularities (e.g., \textit{pick up a cup} $\rightarrow$ \textit{drink} $\rightarrow$ \textit{put down the cup}) or causal chains where certain motions lead to specific outcomes (e.g., a gymnast landing on the ground after a sequence of aerial movements). 
Discriminatively, many actions are difficult to distinguish because they may be semantically related or visually similar while differing in the critical motion itself.
These priors are essential for recognition, but they can also drive systematic hallucination when models over-rely on them instead of visual evidence.
Current Video Large Language Models (VideoLLMs) often fail to decouple these high-level priors from actual visual evidence. 
This lack of deep perception leads to motion hallucinations when models are presented with truncated videos or scenarios where the expected outcome does not occur.

To address these limitations, we introduce \textbf{MoHallBench}, a novel and comprehensive benchmark designed specifically to evaluate motion hallucinations in VideoLLMs. 
As shown in Fig.~\ref{fig:figure1}, MoHallBench assesses motion hallucinations through three axes: \textbf{co-occurrence hallucination}, \textbf{sequential inference hallucination}, and \textbf{similarity hallucination}. 
Co-occurrence hallucination can be further decomposed into object--action, scene--action, and action--action settings, while similarity hallucination is decomposed into semantic and visual similarity. 
For benchmark construction, we compute co-occurrence statistics (conditional probability and pointwise mutual information~\cite{r7pmi}) on object--action, scene--action, and action--action pairs from Charades~\cite{r1charades} to build prior-conflicting negatives in co-occurrence hallucination tasks.
For sequential inference hallucination, we identify action categories with strong process--outcome dependencies, leverage step-level temporal annotations from FineDiving and TAPOS~\cite{r2diving, r3tapos} to separate trigger phases from terminal outcomes, and create adversarial samples by truncating clips before the outcome segment.
For similarity hallucination, we build a three-level semantic hierarchy over 500 atomic actions, and mine visually similar yet motion-different clips.
We further design a bias-aware protocol for binary-choice questions by pairing positive and negative queries for each action instance, yielding metrics that better reflect true capability.

In summary, our contributions are as follows:

\begin{itemize}
    \item We propose \textbf{MoHallBench}, a comprehensive benchmark specifically designed to evaluate motion hallucinations across three axes: co-occurrence, similarity, and sequential inference. It features 11,306 video clips and targets deep spatiotemporal failures through adversarial sampling methods.
    \item We introduce a \textbf{bi-directional questioning protocol}, using both positive and negative framings to mitigate the pervasive affirmation bias. Along with this, we define rigorous metrics \textbf{$\text{Cons}$} and \textbf{$\text{Acc}_{\text{PS/NS}}$} for positive and negative samples to provide a more faithful assessment of a model's true perceptual capability.
    \item Our extensive evaluations reveal a clear decoupling between action recognition and motion-hallucination resistance in current VideoLLMs, with this gap being especially pronounced under sequential inference hallucination axis.
\end{itemize}

\section{Related Work}
\subsection{VideoLLMs}
The capabilities of multimodal large language models are advancing from static image comprehension~\cite{r28blip2, r29instructblip, r26llava} toward the processing of video sequences that contain rich dynamic information~\cite{r37flamingo, r38videochatgpt, r39videollama, r40valley}. Recent innovations in model architectures and video encoding paradigms aim to bridge the representation gap between these modalities. The Qwen3-VL series~\cite{r9qwen3vl} uses an enhanced interleaved-MRoPE mechanism for spatiotemporal modeling, which allows the model to process temporal and spatial positional relationships within mixed contexts of images and videos. Similarly, InternVL3~\cite{r17internvl3} introduces Variable Visual Position Encoding, often abbreviated as V2PE, to support extended multimodal contexts. To optimize efficiency, MiniCPM-V 4.5~\cite{r18minicpm} employs a unified 3D-Resampler architecture that uses a single encoder for both images and videos while compressing multiple frames into a minimal number of visual tokens. VideoLLaMA 3~\cite{r13vlm3} prunes redundant video tokens based on inter-frame differences. 

\begin{figure*}[!t]
    \centering
    \includegraphics[width=\textwidth]{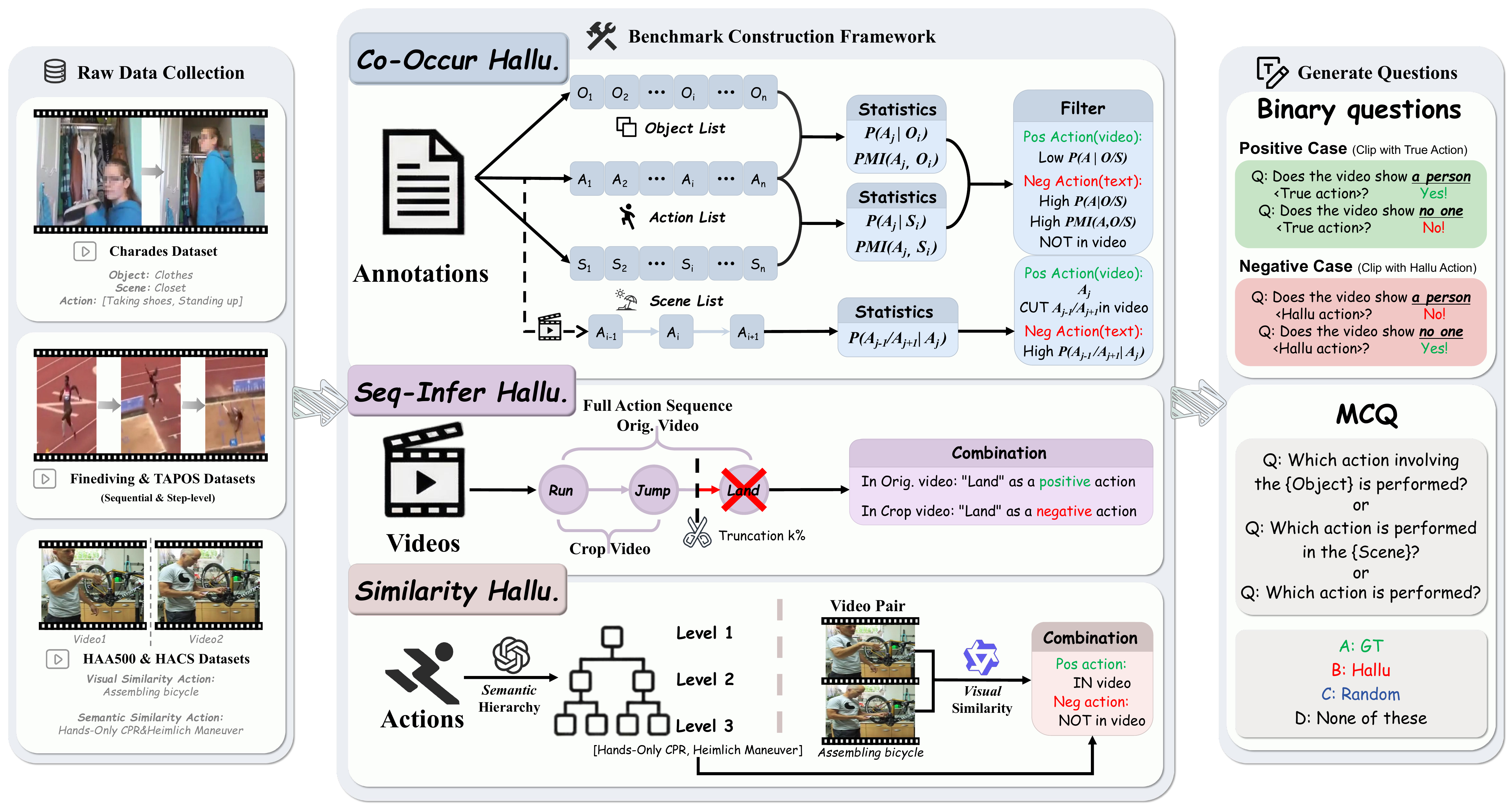}
    \caption{Overview of the MoHallBench pipeline. We curate adversarial samples from multi-source videos along three axes, including co-occurrence hallucination (CH), sequential inference hallucination (SIH), and similarity hallucination (SH), and evaluate VideoLLMs with binary-choice (BC), multiple-choice (MC), and generative protocols using bias-aware metrics.}
    \label{fig:figure2}
\end{figure*}

\subsection{Hallucination in VideoLLMs}
Multimodal hallucination is defined as the generation of plausible descriptions that are inconsistent with the provided visual evidence~\cite{r24hallusurvey, r25hallusurvey2}. 
This issue was first documented in image-based models where models frequently hallucinated nonexistent objects based on co-occurrence patterns in the training data~\cite{r19pope, r31amber, r41chair}. 
As research extends to the video domain, new benchmarks have emerged to address temporal inconsistencies. 
EventHallusion~\cite{r21eventhallusioin} evaluates hallucinations caused by language priors and vision-language correlation biases in rare events. 
VideoHallucer~\cite{r22videohallucier} categorizes hallucinations into intrinsic and extrinsic types, enabling evaluation of both contradictions to video content and unverifiable additions beyond it. 
VidHallu~\cite{r23videhalluc} further constructs visually distinct yet semantically similar video pairs to assess hallucinations in action, temporal sequence, and scene transition understanding. 

Among benchmarks specifically designed for action hallucination, MHBench~\cite{r20mhbench} compares real action videos with videos containing semantically opposite or incomplete actions. 
UNSCENE~\cite{r8mash}, in contrast, uses GPT-4-turbo to construct seemingly plausible action labels under given scenes, and evaluates action-scene hallucination using scene-only and full-context videos.
However, these benchmarks do not systematically examine how action-related spatiotemporal elements are connected to actions, nor do they clearly identify the factors that cause hallucinations.
MoHallBench addresses this limitation by systematically investigating how co-occurrence priors, fine-grained similarities, and sequential inference expectations induce motion hallucinations.

\section{MoHallBench}
We present MoHallBench, a diagnostic benchmark for motion hallucination in VideoLLMs. 
As shown in Fig.~\ref{fig:figure2}, the benchmark is built around three complementary axes: co-occurrence hallucination (CH), sequential inference hallucination (SIH), and similarity hallucination (SH). 
It is evaluated with binary-choice (BC), multiple-choice (MC), and generative protocols, together with bias-aware metrics for rigorous analysis.


\subsection{Co-occurrence Hallucination} 
Co-occurrence hallucination occurs when a model prioritizes learned frequency distributions and high-level priors over the actual visual evidence.
To probe such prior-driven hallucinations, we systematically quantify the co-occurrence frequencies of actions with objects, scenes, and other actions within the Charades dataset~\cite{r1charades}.

\paragraph{Object-Action Hallucination (OAH)} 
This task evaluates the degree to which models rely on statistical interactions between objects and verbs rather than authentic motion perception. We define an action as a tuple $a=(o, v)$ consisting of an interacting object $o$ and a verb $v$, and conduct the analysis over 38 objects and 33 verbs. To quantify the conditioned default prior, we calculate the conditional probability $P$ of a verb given an object: 
\begin{equation} 
P(v \mid o) = \frac{N(o, v)}{\sum_{v'} N(o, v')} 
\end{equation} 
However, a high $P(v \mid o)$ does not necessarily imply a unique association between object $o$ and a verb $v$, as it may be driven by globally common verbs such as \textit{hold} or \textit{take}, which already have high marginal probabilities $P(v)$. To distinguish object-specific frequent interactions from globally frequent ones, we incorporate Pointwise Mutual Information (PMI)~\cite{r7pmi}: 
\begin{equation} 
\text{PMI}(o, v) = \log\frac{P(v \mid o)}{P(v)} 
\end{equation} 
A positive PMI indicates that verb $v$ is more likely to occur in the presence of object $o$ than expected from its global frequency, reflecting a positive object-conditioned prior. In our construction pipeline, positive samples are drawn from clips where the ground-truth action represents a non-default interaction, defined as those within the bottom 50\% of the frequency distribution for a specific object. Negative samples are identified as hard prior distractors that exhibit high conditional probability and positive PMI but are factually absent from the video. In BC tasks, we query the existence of positive and negative actions.
For MC evaluation, the question template is:

\textbf{\textit{``Which action involving the {$o$} is performed in the video?''}}\\
The candidate set includes the ground-truth action, a hard prior distractor, a random action involving the same object but absent from the video, and ``None of these.''

\paragraph{Scene-Action Hallucination (SAH)} 
We extend our investigation to hallucinations induced by the environmental context, defined as the scene $s$, covering 15 scene categories. This task measures whether VideoLLMs prioritize scene-driven common sense over visual evidence when predicting actions. We define the scene-conditioned action frequency as: 
\begin{equation} 
P(a \mid s) = \frac{N(s, a)}{\sum_{a'} N(s, a')} 
\end{equation} 
Similar to the OAH task, we use PMI to distinguish between globally frequent actions and those with a genuine functional relationship to the scene context: 
\begin{equation} 
\text{PMI}(s, a) = \log\frac{P(a \mid s)}{P(a)} 
\end{equation} 
The positive PMI value indicates that the scene provides a strong contextual cue for the action, thereby increasing the likelihood of inducing a hallucination. 
Adversarial scenarios are constructed by pairing a video clip with a negative action that possesses a high $P(a \mid s)$ and a strictly positive PMI value relative to the depicted scene. 
In BC tasks, the model must verify the existence of both the ground-truth action and the hallucinated action in separate queries. 
The MC format uses the following question template:

\textbf{\textit{``Which action is performed in the {$s$} scene shown in the video?''}}\\
The four options include the ground-truth action, a hallucinated action, a randomly chosen action, and ``None of these.'' This setup forces the model to choose between scene-driven common sense and authentic motion perception.
 
\paragraph{Action-Action Hallucination (AAH)} 
This task investigates whether VideoLLMs hallucinate the presence of actions based on strong temporal neighbor priors, with transition statistics computed over 151 actions. In structured activities, certain actions exhibit strong transition probabilities, leading models to over-rely on event scripts rather than temporal evidence. To quantify these associations, we calculate the conditional probabilities for both predecessors and successors within action sequences: 
\begin{equation}
P(a_{i-1} \mid a_i) = \frac{N(a_{i-1}, a_i)}{\sum_{a'} N(a', a_i)}
\end{equation}
\begin{equation}
P(a_{i+1} \mid a_i) = \frac{N(a_i, a_{i+1})}{\sum_{a'} N(a_i, a')}
\end{equation}
where $a_{i-1}$ and $a_{i+1}$ represent the preceding and subsequent actions relative to the target $a_i$, respectively. For adversarial construction, we select the top two most frequent transition patterns for each action category. While positive samples retain $a_i$ in context, negative samples are generated by truncating the video to physically omit expected predecessors or successors. 
In BC tasks, we query the presence of $a_i$ in the original clip and the omitted action in the truncated clip to assess the model's capacity for counterfactual negation. 
In the MC format, the question template is:

\textbf{\textit{``Which action is performed in the video?''}}\\
The options include the ground-truth action, a prior-driven hallucination action, a random distractor sampled from the top five most frequent transitions, and ``None of these.''

\subsection{Sequential Inference Hallucination} 
Sequential inference hallucination, also characterized as over imagination, occurs when a model completes an event script based on process--outcome necessity rather than actual visual evidence. In many structured human activities, a specific initial phase $a_i$ physically or semantically necessitates a terminal result $a_n$. Models often fall into the trap of assuming the sequence has reached its logical conclusion even when the visual evidence is omitted in the video. We curate a subset of 11 motion categories with highly predictable endings, such as \textit{diving}, \textit{jumping}, and \textit{throwing}. 
We manually identify and verify the terminal sub-actions (e.g., \textit{landing} in a long jump) for each action category. 
To construct adversarial samples, we leverage the step-level boundary annotations from source datasets~\cite{r3tapos, r2diving} to precisely isolate the trigger phases from the terminal outcomes, then remove all frames from that point onward. 
In BC tasks, we ask the model to verify whether the terminal action is present in both the original and truncated clips.
For the generative task, we use the prompt:

\textbf{\textit{``Describe the sequence of actions performed by the person in the video.''}}
to elicit a caption for the truncated clip and assess whether the model hallucinates terminal actions that are absent from the observed video.

\subsection{Similarity Hallucination}
Similarity-induced hallucination occurs when models lack the fine-grained discriminative ability needed to distinguish between actions that are either semantically related or visually similar at the surface level.
\paragraph{Semantic Similarity Hallucination (SSH)} 
This task evaluates whether models can distinguish between actions that share high-level semantic purposes but involve distinct physical movements. 
We employ GPT-5.4 Thinking~\cite{r36gpt5} to organize 500 atomic actions into a fine-grained semantic hierarchy. This taxonomy consists of three to four levels of nested semantics, exemplified by the path: \textit{ Social Interaction \& Communication $\rightarrow$ Greeting \& Affection Gestures $\rightarrow$ Respectful Greeting Postures}. 
To construct adversarial pairs, we identify actions within the same fine-grained leaf cluster, such as \textit{Cardiopulmonary Resuscitation (CPR)} and the \textit{Heimlich maneuver}, which are semantically related but visually different. For binary choice tasks, we present a video of a specific action $a_{\text{true}}$ and query the model regarding the presence of its semantic distractor $a_{\text{hallu}}$. 
In the MC format, the question template is:

\textbf{\textit{``Which action is performed in the video?''}}\\
The options include the ground-truth action, a semantic distractor, an unrelated action from a different cluster, and ``None of these.''

\paragraph{Visual Similarity Hallucination (VSH)} 
We further investigate hallucinations triggered by visual resemblance where the background, subject, and interacting objects are nearly identical, but the target motion is absent. 
We employ a discriminative reranker model, Qwen3--VL--Reranker~\cite{r6ranker}, to identify negative clips from the same source video that exhibit high visual similarity to the positive clip but lack the labeled action. 
After manual verification, we select pairs that exhibit the highest visual similarity but possess different motion content. For binary evaluation, the model is asked to verify the presence of the target action in both positive and visually similar negative clips. 
In the dual-video MC task, we present either a positive or a negative clip and prompt the model with:

\textbf{\textit{``Which action is performed in the video?''}}\\
The model is asked to choose from four options: the target action, two random actions, and ``None of these'' 
The expected answer is the target action for the positive clip and "None of these" for the negative one.

\subsection{Dataset Statistics}
\begin{figure}[t]
    \centering
    \includegraphics[width=\linewidth]{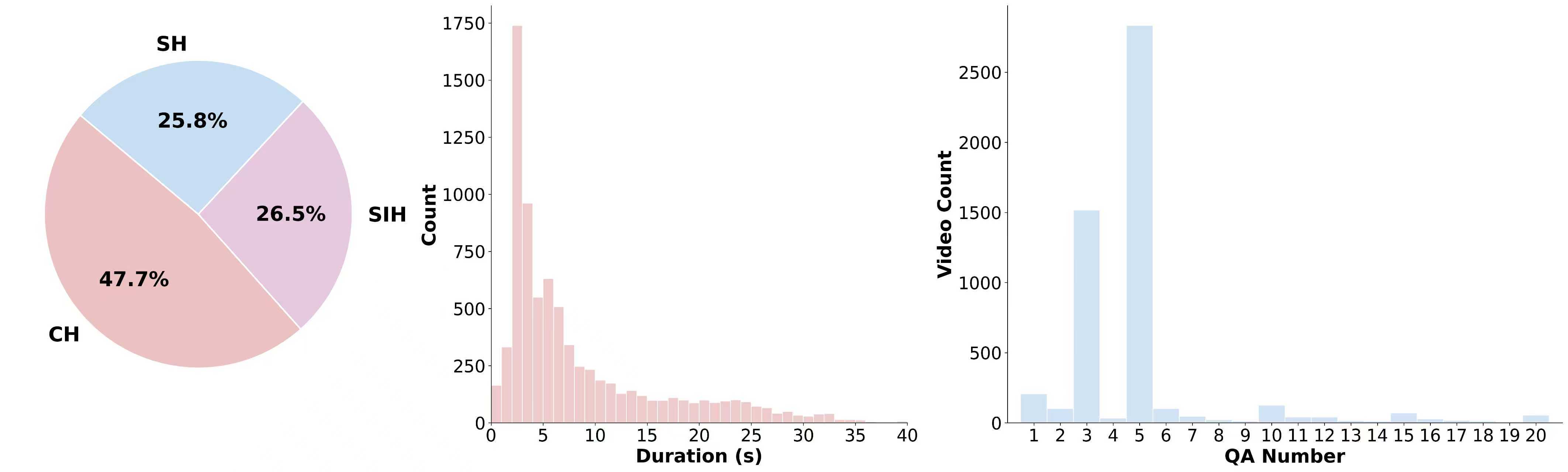}
    \caption{Dataset statistics of MoHallBench. From left to right: category distribution across CH, SIH, and SH; the distribution of QA pairs per video; and the clip duration distribution.}
    \label{fig:stats}
\end{figure}

MoHallBench is curated from Charades~\cite{r1charades}, HAA500~\cite{r4haa500}, HACS~\cite{r5hacs}, FineDiving~\cite{r2diving}, and TAPOS~\cite{r3tapos}. 
It comprises 11,306 video clips and 40,493 question-answer pairs, with an average of 3.58 questions per video and a mean clip duration of 8.29 seconds.
This temporal scale captures motion signatures effectively while minimizing context redundancy.
The distributions of categories, question-answer (QA) pairs per video, and duration are illustrated in Fig.~\ref{fig:stats}.
The dataset maintains a 1:1 ratio between positive and negative samples for binary tasks, and for multiple-choice questions, the four candidate options are randomly shuffled to eliminate positional bias.

\subsection{Evaluation Metrics} 
We evaluate on three task formats, including binary choice (BC), multiple-choice (MC), and open-ended generative questions. 
For binary tasks, existing benchmarks often suffer from a systemic yes-bias problem, where models tend to provide affirmative responses regardless of the visual evidence~\cite{r34VQA, r35VQAv2}. 
To address this, we implement a bi-directional questioning protocol. For each action $a$, the model is evaluated with two complementary queries:

\textbf{Positive query: \textit{``Does the video show $a$?''}}

\textbf{Negative query: \textit{``Does the video show no one $a$?''}}\\
Based on this protocol, we denote the raw accuracy scores as $a_{\text{pos/neg}}^{+/-}$, where the subscript indicates the question type and the superscript indicates the ground-truth label.
The debiased accuracy metrics, $\text{Acc}_{\text{PS}}$ and $\text{Acc}_{\text{NS}}$, can effectively eliminate the bias by averaging performance over flipped ground-truth labels:

\begin{equation}
    \text{Acc}_{\text{PS}} = \frac{1}{2}(a_{\text{pos}}^{+} + a_{\text{neg}}^{+}), \quad \text{Acc}_{\text{NS}} = \frac{1}{2}(a_{\text{pos}}^{-} + a_{\text{neg}}^{-})
\end{equation}
To impose a stricter evaluation of pairwise discrimination, we introduce $\text{Q-PairAcc}$, which counts a successful prediction only if the model correctly answers both the positive and negative queries within an adversarial pair:

\begin{equation}
    \text{Q-PairAcc} = \frac{1}{N} \sum_{i=1}^{N} \mathbb{I} \left[ \hat{y}_{i,\text{pos}} = y_{i,\text{pos}} \wedge \hat{y}_{i,\text{neg}} = y_{i,\text{neg}} \right]
\end{equation}
where $y, \hat{y} \in \{0, 1\}$ represent the ground-truth and predicted labels, respectively.
Furthermore, to assess the internal logical robustness, we define the consistency metric $\text{Cons}$ as the ratio of complementary answers to opposing framings: 
\begin{equation} 
    \text{Cons} = \frac{1}{N} \sum_{i=1}^{N} \mathbb{I} \left[ \hat{y}_{i,\text{neg}} = 1 - \hat{y}_{i,\text{pos}} \right]
\end{equation}
A significantly low $\text{Cons}$ indicates that the model is heavily influenced by both the yes-bias and query-framing bias, rendering the debiased accuracy metrics unreliable.

For MC tasks, we report standard accuracy.  
For generative tasks in SIH, we use DeepSeek~\cite{r32ds} together with manual review to detect whether generated descriptions erroneously entail omitted outcome actions and define $\text{Hallu Rate}$ as the proportion of descriptions containing these hallucinated outcomes.

\begin{table*}[!t]
\centering
\scriptsize
\renewcommand{\arraystretch}{1.10}
\setlength{\tabcolsep}{3pt}

\caption{\textbf{Overall results on MoHallBench.}
For BC, we report positive-sample accuracy (AP, $\text{Acc}_{\text{PS}}$),
negative-sample accuracy (AN, $\text{Acc}_{\text{NS}}$), consistency
(CN, Cons), and pair accuracy (QP, Q-PairAcc).
For auxiliary tasks, we report multiple-choice accuracy (MA) for CH and SH,
and hallucination rate (HR, $\downarrow$) for SIH. All values are reported as percentages.}
\label{tab:overall_motionhallubench_compact}

\resizebox{\textwidth}{!}{%
\begin{tabular}{l*{15}{c}}
\toprule
\multirow{3}{*}{\textbf{Model}}
& \multicolumn{15}{c}{\textbf{Co-occurrence Hallucination (CH)}} \\
\cmidrule(lr){2-16}
& \multicolumn{5}{c}{\textbf{OAH}}
& \multicolumn{5}{c}{\textbf{SAH}}
& \multicolumn{5}{c}{\textbf{AAH}} \\
\cmidrule(lr){2-6}
\cmidrule(lr){7-11}
\cmidrule(lr){12-16}
& \textbf{AP} & \textbf{AN} & \textbf{CN} & \textbf{QP} & \textbf{MA}
& \textbf{AP} & \textbf{AN} & \textbf{CN} & \textbf{QP} & \textbf{MA}
& \textbf{AP} & \textbf{AN} & \textbf{CN} & \textbf{QP} & \textbf{MA} \\
\midrule
InternVL3-8B
& 83.6 & 38.7 & \textbf{73.5} & \textbf{27.9} & 63.8
& \textbf{91.7} & \textbf{55.5} & \textbf{82.5} & \textbf{50.1} & 74.5
& \textbf{89.0} & 23.5 & \textbf{76.7} & \textbf{19.9} & \textbf{72.6} \\

LLaVA-Video-7B-Qwen2
& 84.5 & 30.0 & 63.2 & 19.3 & \textbf{47.6}
& 89.1 & 46.5 & 66.3 & 37.9 & \textbf{77.5}
& 85.6 & 26.5 & 64.9 & 19.1 & 64.0 \\

MiniCPM-V-4.5
& 70.3 & \textbf{41.5} & 37.6 & 18.4 & 48.4
& 77.9 & 49.8 & 53.5 & 34.3 & 56.4
& 72.8 & 30.3 & 47.5 & 15.1 & 63.7 \\

Molmo-2-8B
& 85.0 & 40.7 & 54.8 & 28.9 & \textbf{75.0}
& 91.2 & 57.0 & 65.2 & 49.9 & 78.2
& 88.0 & 30.5 & 68.3 & 24.4 & 72.1 \\

Perception-LM-8B
& 79.8 & \textbf{50.8} & 68.7 & \textbf{36.8} & 73.6
& 86.1 & \textbf{73.3} & 78.2 & \textbf{63.1} & \textbf{83.5}
& 87.3 & \textbf{41.9} & 71.9 & \textbf{34.8} & \textbf{80.8} \\

Qwen2.5-VL-7B-Instruct
& 76.2 & 38.1 & 38.3 & 17.3 & 60.8
& 80.3 & 43.1 & 37.6 & 26.3 & 72.8
& 72.9 & \textbf{35.3} & 39.3 & 14.3 & 65.2 \\

Qwen2-VL-7B-Instruct
& \textbf{85.7} & 28.0 & 59.6 & 18.6 & 69.4
& 90.6 & 39.2 & 62.0 & 32.2 & 71.0
& 88.8 & 20.7 & 71.0 & 15.5 & 72.3 \\

Qwen3-VL-8B-Instruct
& 74.5 & 50.5 & 70.0 & 33.1 & 52.0
& 88.6 & 61.3 & \textbf{85.0} & 53.6 & 76.1
& 81.2 & 34.5 & 74.7 & 25.7 & 70.0 \\

Tarsier2-7B
& 84.0 & 35.8 & 59.0 & 24.5 & 73.7
& 90.9 & 56.0 & 64.7 & 48.5 & 82.4
& 87.1 & 34.5 & 62.0 & 26.9 & \textbf{80.8} \\

VideoLlama3-7B
& \textbf{89.9} & 29.5 & \textbf{78.1} & 24.7 & 62.1
& \textbf{92.0} & 52.0 & 82.4 & 47.1 & 53.8
& \textbf{91.9} & 28.0 & \textbf{79.0} & 25.0 & 63.1 \\
\bottomrule
\end{tabular}%
}

\vspace{0.9em}

\resizebox{\textwidth}{!}{%
\begin{tabular}{l*{15}{c}}
\toprule
\multirow{3}{*}{\textbf{Model}}
& \multicolumn{5}{c}{\textbf{Sequential Inference Hallucination (SIH)}}
& \multicolumn{10}{c}{\textbf{Similarity Hallucination (SH)}} \\
\cmidrule(lr){2-6}
\cmidrule(lr){7-16}
& \multicolumn{5}{c}{\textbf{SIH}}
& \multicolumn{5}{c}{\textbf{SSH}}
& \multicolumn{5}{c}{\textbf{VSH}} \\
\cmidrule(lr){2-6}
\cmidrule(lr){7-11}
\cmidrule(lr){12-16}
& \textbf{AP} & \textbf{AN} & \textbf{CN} & \textbf{QP} & \textbf{HR}$\downarrow$
& \textbf{AP} & \textbf{AN} & \textbf{CN} & \textbf{QP} & \textbf{MA}
& \textbf{AP} & \textbf{AN} & \textbf{CN} & \textbf{QP} & \textbf{MA} \\
\midrule
InternVL3-8B
& \textbf{99.8} & 5.6 & \textbf{94.5} & 5.5 & 66.2
& \textbf{95.4} & 27.7 & \textbf{83.5} & 25.8 & 78.9
& \textbf{95.5} & 30.8 & \textbf{83.8} & 27.9 & 58.3 \\

LLaVA-Video-7B-Qwen2
& 99.7 & 7.2 & 92.6 & 6.9 & 36.0
& 93.6 & 25.8 & 70.3 & 21.2 & 77.6
& 94.7 & 28.7 & 71.2 & 24.6 & 56.0 \\

MiniCPM-V-4.5
& 56.9 & \textbf{48.7} & 9.1 & 6.8 & 54.0
& 88.8 & 27.9 & 69.4 & 23.4 & 74.1
& 92.5 & 29.3 & 72.2 & 24.6 & \textbf{64.1} \\

Molmo-2-8B
& 98.0 & 12.9 & 85.2 & 11.1 & 43.3
& 93.8 & 30.7 & 72.0 & 27.0 & 77.3
& 93.5 & 29.3 & 70.1 & 24.4 & 51.2 \\

Perception-LM-8B
& 83.4 & 37.4 & 63.5 & 22.3 & 53.7
& 86.9 & 46.1 & 70.5 & 38.1 & 84.8
& 91.6 & 41.1 & 78.2 & 34.5 & 55.1 \\

Qwen2.5-VL-7B-Instruct
& 96.3 & 14.0 & 82.4 & 11.0 & 56.8
& 77.4 & 40.0 & 37.6 & 20.7 & 79.3
& 86.6 & 33.9 & 53.5 & 22.3 & 56.6 \\

Qwen2-VL-7B-Instruct
& 64.1 & 42.3 & 21.9 & 7.1 & 66.8
& 92.1 & 20.6 & 73.8 & 16.4 & 78.7
& 94.3 & 21.7 & 73.7 & 17.4 & 52.0 \\

Qwen3-VL-8B-Instruct
& 95.7 & 14.5 & 85.0 & 11.5 & 57.0
& 88.0 & \textbf{46.2} & 81.6 & \textbf{39.3} & 81.0
& 91.5 & \textbf{42.6} & 82.1 & \textbf{36.2} & 59.2 \\

Tarsier2-7B
& 98.4 & 30.1 & 78.9 & \textbf{29.1} & \textbf{5.7}
& 94.3 & 30.6 & 74.0 & 27.6 & \textbf{87.5}
& 93.3 & 36.4 & 70.8 & 31.7 & 54.4 \\

VideoLlama3-7B
& 95.7 & 19.4 & 82.3 & 18.1 & 52.4
& 94.3 & 24.3 & 83.2 & 22.7 & 72.5
& 94.6 & 26.2 & 81.9 & 23.6 & 60.5 \\
\bottomrule
\end{tabular}%
}
\end{table*}


\section{Experiment}

\subsection{Experimental Setup}
\textbf{Baselines.} We assess the motion hallucination behavior of ten recent representative open-source VideoLLMs on MoHallBench. 
The selected baselines include Qwen3-VL-8B-Instruct~\cite{r9qwen3vl}, Molmo-2-8B~\cite{r12molmo2}, Perception-LM-8B~\cite{r16plm}, InternVL3-8B~\cite{r17internvl3}, VideoLlama3-7B~\cite{r13vlm3}, Qwen2-VL-7B~\cite{r10qwen2vl}, Tarsier2-7B~\cite{r14tarsier2}, LLaVA-Video-7B-Qwen2~\cite{r15lnq}, Qwen2.5-VL-7B~\cite{r11qwen2.5vl}, and MiniCPM-V-4.5~\cite{r18minicpm}. 
To ensure a fair comparison, we evaluate all models at a comparable 7B--8B parameter scale.

\noindent\textbf{Implementation Details.} We adopt a uniform sampling strategy to extract 32 frames from each video clip. 
This frame count is selected because empirical evidence suggests that model performance typically reaches a peak within the 16 to 32 frame range and does not improve significantly with further increases.
Inference is conducted on NVIDIA RTX 4090 hardware. 
To eliminate stochastic variance and produce deterministic outputs, we set the decoding temperature to zero and deactivate top-p sampling.

\subsection{Evaluating on MoHallBench}

Tab.~\ref{tab:overall_motionhallubench_compact} presents the evaluation results on MoHallBench, indicating that motion hallucination is a pervasive issue across all evaluated VideoLLMs. 
A fundamental observation is the decoupling between action recognition and motion falsification. While most models achieve high scores in positive accuracy, their performance drops significantly when faced with adversarial negative samples. 
For example, InternVL3 achieves 83.6\% in $\text{Acc}_{\text{PS}}$ of the OAH task but only 38.7\% in $\text{Acc}_{\text{NS}}$, which results in a low $\text{Q-PairAcc}$ of 27.9\%.
Furthermore, the consistency metrics show that model performance remains unsatisfactory even when response bias is minimal, indicating that the observed errors cannot be primarily attributed to response bias and instead reflect genuine failures in resisting motion hallucination.

\noindent\textbf{CH.} In the CH category, models perform better on scene-action tasks compared to object-action or action-action tasks. Perception-LM stands out as the top performer in the SAH category with a pairwise accuracy of 63.1\%, which is significantly higher than other baselines. This performance gap suggests that specialized perception models may possess more robust spatiotemporal feature extraction capabilities.

\noindent\textbf{SIH.}
Among the different hallucination categories, SIH represents the most significant challenge. 
Models exhibit an extreme tendency to over-infer the completion of a motion sequence. 
Even the top-performing Tarsier2~\cite{r14tarsier2} only manages a $\text{Q-PairAcc}$ of 29.1\% in this category, highlighting the significant challenge of maintaining temporal faithfulness.
This result confirms that current architectures struggle to distinguish between the initiation and the actual execution of a movement.

\noindent\textbf{SH.} Regarding SH, almost all models exhibit a striking asymmetry, characterized by exceptionally high $\text{Acc}_{\text{PS}}$ but dismal $\text{Acc}_{\text{NS}}$. 
This significant divergence underscores their profound vulnerability to semantic or visual similarity-induced negative samples.
Even the top-performing Qwen3~\cite{r9qwen3vl} only achieves a peak Q-PairAcc of 39.3\% and 36.2\% across the two similarity categories.


\subsection{Analysis of the Causes of Hallucinations}
In this section, we investigate the underlying factors that trigger motion hallucinations by conducting a series of diagnostic experiments using Qwen3-VL-8B-Instruct~\cite{r9qwen3vl}. 
All analyses are conducted on the BC task, where we examine negative samples with different levels of hallucination inducement and analyze how hallucination severity changes accordingly. 
The resulting quantitative trends not only reveal the factors underlying the observed hallucinations but also confirm the validity of our benchmark settings. 

\begin{table}[!t]
\centering
\footnotesize
\renewcommand{\arraystretch}{1.12}
\setlength{\tabcolsep}{3.5pt}
\caption{Equal-width binning results for co-occurrence hallucination (CH) under Qwen3-VL in the binary-choice (BC) setting. We report positive-query accuracy (posq), negative-query accuracy (negq), and $\text{Acc}_{\text{NS}}$.}
\label{tab:ch_binning_qwen3}
\resizebox{\columnwidth}{!}{%
\begin{tabular}{llccc}
\toprule
\textbf{Subtype} & \textbf{Bin Range} & \textbf{posq (\%)} & \textbf{negq (\%)} & \textbf{$\text{Acc}_{\text{NS}}$ (\%)} \\
\midrule
\multirow{4}{*}{OA} & $(0.0454, 0.2770]$ & 70.8 & 43.8 & 57.3 \\
& $(0.2770, 0.5080]$ & 60.4 & 50.0 & 55.2 \\
& $(0.5080, 0.7380]$ & 33.3 & 16.7 & 25.0 \\
& $(0.7380, 0.9680]$ & 27.1 & 18.8 & 22.9 \\
\midrule
\multirow{4}{*}{SA} & $[0.0001, 0.0157]$ & 89.4 & 43.0 & 66.2 \\
& $(0.0157, 0.0313]$ & 68.9 & 43.7 & 56.3 \\
& $(0.0313, 0.0469]$ & 62.6 & 32.6 & 47.6 \\
& $(0.0469, 0.0625]$ & 47.6 & 49.6 & 48.6 \\
\midrule
\multirow{4}{*}{AA} & $[0.0247, 0.1045]$ & 47.1 & 34.5 & 40.8 \\
& $(0.1045, 0.1843]$ & 34.5 & 10.3 & 22.4 \\
& $(0.1843, 0.2641]$ & 36.8 & 9.2 & 23.0 \\
& $(0.2641, 0.3439]$ & 28.7 & 8.1 & 18.4 \\
\bottomrule
\end{tabular}
}
\end{table}

\begin{figure*}[!t]
    \centering
    \includegraphics[width=\textwidth]{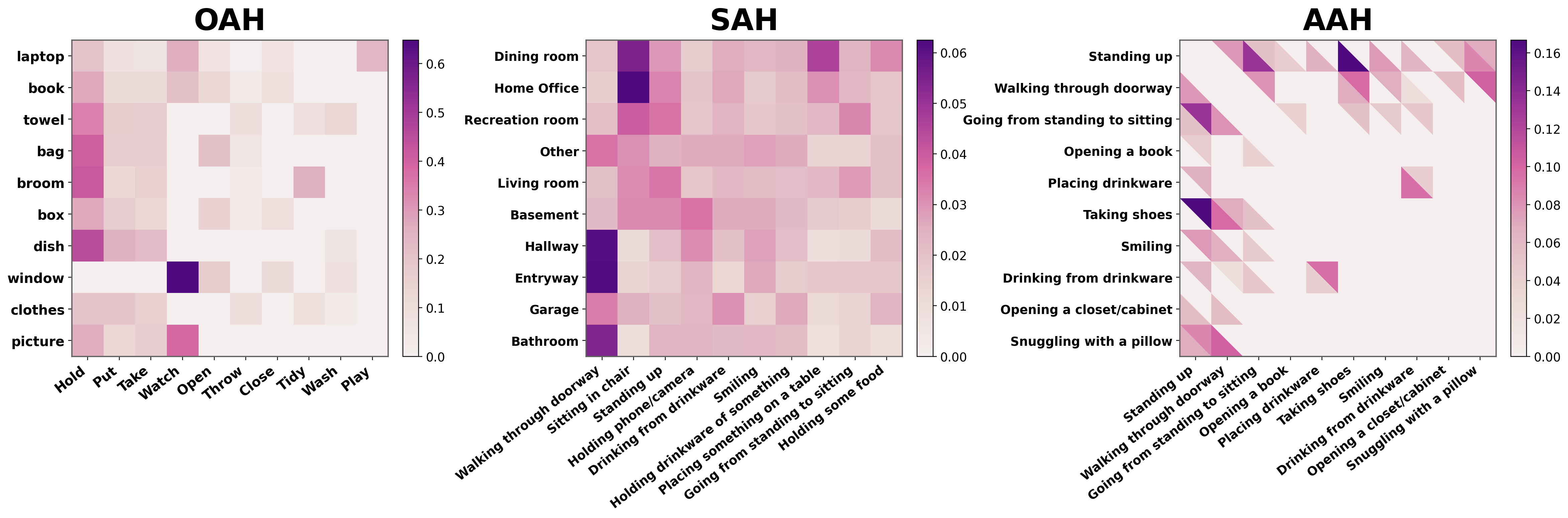}
    \caption{Top-10 co-occurrence structures for OAH, SAH, and AAH. For AAH, the horizontal-axis action is denoted by $x$ and the vertical-axis action is denoted by $y$; the lower triangle denotes $P(y\mid x)$ and the upper triangle denotes $P(x\mid y)$.}
    \label{fig:cooccurrence}
\end{figure*}

\noindent
\textbf{CH.} Fig.~\ref{fig:cooccurrence} visualizes the top-10 co-occurrence patterns for OAH, SAH, and AAH, revealing clear prior regularities in object--action, scene--action, and action--action relations. We then examine how hallucination severity changes with context-conditioned co-occurrence strength. 
For each CH subtype, we apply equal-width binning over the corresponding conditional probability $P$ (sorted from low to high) and enforce balanced sample counts across bins by downsampling to the minimum bin size.
For object-action pairs (OA), accuracy consistently degrades as $P(v\mid o)$ increases. In the lowest-probability bin, the model achieves $57.29\%$ $\text{Acc}_{\text{NS}}$, but performance drops to $22.92\%$ in the highest-probability bin. 
A similar monotonic pattern appears in scene-action pairs (SA) and action-action pairs (AA), with $\text{Acc}_{\text{NS}}$ decreasing from $66.20\%$ to $48.61\%$ and $40.80\%$ to $18.39\%$ as transition probability increases.

\begin{table}[t]
\centering
\footnotesize
\caption{High-$P$ object--action (OA) and scene--action (SA) subsets stratified by PMI bands under Qwen3-VL in the binary-choice (BC) setting.}
\label{tab:pmi_band_qwen3}
\resizebox{\columnwidth}{!}{%
\begin{tabular}{llccc}
\toprule
\textbf{Subtype} & \textbf{Band} & \textbf{posq (\%)} & \textbf{negq (\%)} & \textbf{$\text{Acc}_{\text{NS}}$ (\%)} \\
\midrule
\multirow{3}{*}{OA} & P\_HIGH / PMI\_HIGH & 73.5 & 61.1 & 67.3 \\
& P\_HIGH / PMI\_MID   & 35.2 & 8.6 & 21.9 \\
& P\_HIGH / PMI\_LOW   & 41.0 & 19.7 & 30.3 \\
\midrule
\multirow{3}{*}{SA} & P\_HIGH / PMI\_HIGH & 89.3 & 43.8 & 66.6 \\
& P\_HIGH / PMI\_MID   & 75.8 & 45.5 & 60.6 \\
& P\_HIGH / PMI\_LOW   & 56.7 & 26.7 & 41.7 \\
\bottomrule
\end{tabular}
}
\end{table}

\noindent\textbf{The influence of PMI.}
High conditional probability can arise from two distinct sources: globally frequent default actions and context-specific functional actions. To disentangle these two effects, we further partition the high-$P$ subsets in OA and SA into low-, mid-, and high-PMI bands.

In OAH, the pair accuracy is highest in the high-PMI band ($0.673$), but drops sharply in the medium-PMI band ($0.219$) and remains low in the low-PMI band ($0.303$). 
A similar trend is observed in SAH. This reveals a counterintuitive finding that even under high $P$, lower-PMI actions are more hallucination-prone.

\begin{figure*}[t]
    \centering
    \includegraphics[width=\textwidth]{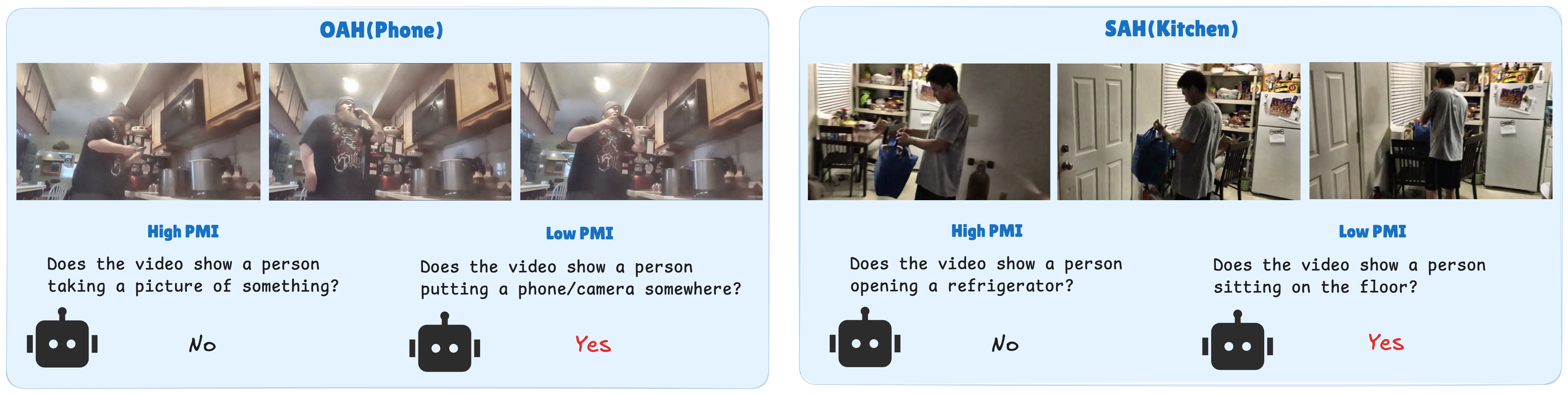}
    \caption{Performance stratified by PMI in high-$P$ regions. Lower-PMI bands show more severe hallucination under both OA and SA settings.}
    \label{fig:pmi}
\end{figure*}

A plausible explanation is that high-$P$/high-PMI actions are functionally specific to the context and therefore require stronger visual confirmation. In contrast, high-$P$/low-PMI actions (e.g., \textit{pick up}, \textit{put down}, \textit{enter}) are broadly compatible with many contexts but weakly diagnostic, making them easier to hallucinate from priors alone. Figure~\ref{fig:pmi} and our case studies support this interpretation: Qwen3-VL correctly rejects high-PMI distractors but frequently affirms low-PMI default actions that are absent in the video.

\begin{table}[!t]
\centering
\footnotesize
\caption{Sequential inference hallucination (SIH) truncation-depth analysis on diving under Qwen3-VL in the binary-choice (BC) setting.}
\label{tab:sih_truncation_qwen3}
\resizebox{\columnwidth}{!}{%
\begin{tabular}{llccc}
\toprule
\textbf{Phase} & \textbf{Progress (\%)} & \textbf{posq (\%)} & \textbf{negq (\%)} & \textbf{Avg $\text{Acc}_{\text{NS}}$ (\%)} \\
\midrule
\multirow{3}{*}{Takeoff}
& 10.0 & 80.7 & 80.1 & 80.4 \\
& 20.0 & 45.0 & 63.3 & 54.2 \\
& 30.0 & 13.9 & 41.4 & 27.6 \\
\midrule
Transition 
& 40.0 & 3.1 & 24.0 & 13.5 \\
\midrule
\multirow{4}{*}{Aerial Maneuver}
& 50.0 & 0.1 & 11.6 & 5.9 \\
& 60.0 & 0.0 & 4.0 & 2.0 \\
& 70.0 & 0.0 & 0.9 & 0.5 \\
& 80.0 & 0.0 & 1.3 & 0.7 \\
\bottomrule
\end{tabular}
}
\end{table}

\noindent\textbf{SIH.}
For SIH, using diving as a representative case, we examine the trigger-outcome association by truncating videos at different execution stages. 
Diving typically follows the steps of: \textit{Takeoff} $\rightarrow$ \textit{Aerial Maneuver} $\rightarrow$ \textit{Entry}. 
In our annotations, aerial execution dominates around 50\%--70\% progress, while water entry occurs near the end.
As truncation moves later, performance collapses rapidly: average $\text{Acc}_{\text{NS}}$ falls from $80.39\%$ (10\%) to $13.54\%$ (40\%), then to $5.86\%$ (50\%), and approaches zero beyond 60\%. At 70\%, the score is only $0.47\%$.

This trajectory suggests that, once the triggering motion becomes sufficiently salient, the model prematurely infers task completion and predicts an unobserved outcome. 
Overall, SIH exposes a systematic trigger-to-outcome shortcut: instead of verifying whether the terminal sub-action is visually present, the model infers the terminal state from early- or mid-stage cues. Consequently, over-imagination remains severe even when adversarial truncation removes the terminal action from the observed clip.
\\
\noindent\textbf{SH.}
We analyze the relationship between clip-level visual similarity and hallucination. As similarity between the negative clip and the positive reference increases, $\text{Acc}_{\text{NS}}$ drops monotonically from $61.46\%$ to $12.50\%$ across four bins ($[0.0015,0.1739)$ $\rightarrow$ $[0.5189,0.6914]$). In the highest-similarity bin, the model is almost unable to reject absent actions, showing that near-duplicate appearance severely weakens motion-grounded discrimination.

We further probe semantic granularity using a hierarchical action taxonomy. \textit{L1} pairs share the same finest-grained leaf category (original benchmark setting). We then construct less-confusable alternatives: \textit{L2} pairs share the same second-last parent but differ at the leaf level, and \textit{L3} pairs share the same third-last parent but differ at the second-last level. The VSH and SSH analyses are summarized jointly in Tab.~\ref{tab:sh_visual_binning_qwen3}. The results are decisive: $\text{Acc}_{\text{NS}}$ improves from $40.87\%$ (L1) to $83.73\%$ (L2) and $87.59\%$ (L3). Once leaf-level semantic overlap is removed, recognition quality recovers substantially, validating that our L1 construction precisely targets fine-grained semantic confusion rather than generic label ambiguity.


\begin{table}[t]
\centering
\footnotesize
\caption{Similarity hallucination (SH) analysis under Qwen3-VL in the binary-choice (BC) setting, covering visual similarity hallucination (VSH) through visual-similarity bins and semantic similarity hallucination (SSH) through semantic hierarchy levels.}
\label{tab:sh_visual_binning_qwen3}
\resizebox{\columnwidth}{!}{%
\begin{tabular}{llccc}
\toprule
\textbf{Subtype} & \textbf{Bin/Level} & \textbf{posq (\%)} & \textbf{negq (\%)} & \textbf{$\text{Acc}_{\text{NS}}$ (\%)} \\
\midrule
\multirow{4}{*}{VSH} & $(0.0015, 0.1739]$ & 68.8 & 54.2 & 61.5 \\
& $(0.1739, 0.3464]$ & 54.2 & 39.6 & 46.9 \\
& $(0.3464, 0.5189]$ & 41.7 & 22.9 & 32.3 \\
& $(0.5189, 0.6914]$ & 18.8 & 6.2 & 12.5 \\
\midrule
\multirow{3}{*}{SSH} & Original L1 & 44.7 & 37.1 & 40.9 \\
& L2 & 97.6 & 69.8 & 83.7 \\
& L3 & 99.2 & 76.0 & 87.6 \\
\bottomrule
\end{tabular}
}
\end{table}

\section{Conclusion}
In this paper, we studied motion hallucination in VideoLLMs, a video-specific failure mode where models infer human motions that are not supported by visual evidence. To this end, we introduced \textbf{MoHallBench}, a diagnostic benchmark that systematically evaluates motion hallucination from three complementary perspectives: co-occurrence priors, similarity confusion and sequential inference. 
We further proposed a bi-directional questioning protocol together with bias-aware metrics to reduce the influence of affirmation bias and provide a more faithful assessment of motion understanding.
Extensive experiments on ten recent open-source VideoLLMs show that motion hallucination remains a pervasive and severe problem. 
Among all settings, sequential inference hallucination is the most challenging, indicating that current models are especially prone to over-inferring expected outcomes from partial temporal cues. 
Our further analyses also confirm that stronger co-occurrence priors, deeper trigger progression, and higher semantic or visual similarity all amplify hallucination severity.
We hope MoHallBench can serve as a useful benchmark for future research on diagnosing, evaluating, and mitigating motion hallucination in VideoLLMs.

\section{Limitation}
MoHallBench still has several limitations. 
First, MoHallBench mainly focuses on human motion hallucination and does not yet cover broader video hallucination phenomena such as object-state changes or long-horizon event reasoning. 
In addition, some settings rely on curated adversarial construction, which may not fully reflect the diversity of open-world video hallucinations.
Finally, this work focuses on diagnosing motion hallucination rather than mitigating it. We leave the development of mitigation strategies that improve temporal faithfulness, motion-grounded reasoning, and reliable video understanding in open-world scenarios to future work.

\bibliography{aaai2027}

\appendix
\section{Benchmark Construction Details}

\subsection{Co-occurrence Construction Details}
We provide representative statistics for the co-occurrence structures used to construct co-occurrence hallucination (CH), which includes object--action hallucination (OAH), scene--action hallucination (SAH), and action--action hallucination (AAH). These statistics are computed on Charades for two purposes. They verify that our hard negatives are supported by strong contextual priors and show that these priors are highly structured rather than random annotation artifacts. For object--action (OA) and scene--action (SA) pairs, we report both the conditional probability and PMI used in our filtering pipeline. For action--action (AA) pairs, we report directional transition statistics for both the previous-action and next-action settings to capture asymmetric temporal dependencies.

Tab.~\ref{tab:appendix_oa_sa_examples} lists representative OA and SA pairs with high context-conditioned priors. The OA examples are dominated by functionally canonical interactions, such as doorknob $\rightarrow$ grasping, refrigerator $\rightarrow$ opening/closing, and chair $\rightarrow$ sitting. Similarly, the SA examples show that many actions are strongly licensed by scene context, e.g., kitchen $\rightarrow$ cooking, bathroom $\rightarrow$ washing hands, and bedroom $\rightarrow$ lying on a bed. These patterns make them effective hard negatives. They are highly plausible given the visible context, yet absent from the target clip, thereby testing whether the model verifies the motion evidence rather than relying on context alone.

\begin{table*}[t]
\centering
\scriptsize
\caption{Representative object--action and scene--action co-occurrence statistics used in benchmark construction. We show the top-15 examples ranked by contextual relevance after filtering.}
\label{tab:appendix_oa_sa_examples}
\resizebox{\textwidth}{!}{%
\begin{tabular}{cllccc|cllccc}
\toprule
\multicolumn{6}{c|}{\textbf{OA: Representative Object--Action Co-occurrences}} & \multicolumn{6}{c}{\textbf{SA: Representative Scene--Action Co-occurrences}} \\
\textbf{Rank} & \textbf{Object} & \textbf{Action Description} & \textbf{Count} & \textbf{$P(v\mid o)$} & \textbf{PMI} & \textbf{Rank} & \textbf{Scene} & \textbf{Action Description} & \textbf{Count} & \textbf{$P(a\mid s)$} & \textbf{PMI} \\
\midrule
1 & doorknob & Grasping onto a doorknob & 81 & 0.8020 & 6.1599 & 1 & Dining room & Sitting at a table & 235 & 0.0600 & 1.4624 \\
2 & chair & Sitting in a chair & 1591 & 0.9678 & 2.5173 & 2 & Laundry room & Washing some clothes & 83 & 0.0296 & 2.9420 \\
3 & television & Watching television & 461 & 0.8865 & 2.7090 & 3 & Kitchen & Someone is cooking something & 399 & 0.0392 & 1.8399 \\
4 & sofa/couch & Sitting on sofa/couch & 160 & 0.8421 & 2.3782 & 4 & Bathroom & Watching something in a mirror & 159 & 0.0407 & 1.5364 \\
5 & window & Watching/looking outside of a window & 393 & 0.6485 & 2.3964 & 5 & Living room & Sitting on sofa/couch & 451 & 0.0426 & 1.4383 \\
6 & mirror & Watching something/someone/themselves in a mirror & 583 & 0.6478 & 2.3952 & 6 & Pantry & Putting something on a shelf & 80 & 0.0393 & 1.3806 \\
7 & refrigerator & Opening a refrigerator & 306 & 0.5635 & 2.2342 & 7 & Pantry & Taking food from somewhere & 88 & 0.0432 & 1.1305 \\
8 & refrigerator & Closing a refrigerator & 237 & 0.4365 & 2.4438 & 8 & Laundry room & Putting clothes somewhere & 116 & 0.0414 & 1.1771 \\
9 & door & Opening a door & 1057 & 0.5465 & 2.2036 & 9 & Kitchen & Opening a refrigerator & 272 & 0.0267 & 1.7525 \\
10 & door & Closing a door & 822 & 0.4250 & 2.4172 & 10 & Kitchen & Closing a refrigerator & 206 & 0.0202 & 1.7366 \\
11 & cup/glass/bottle & Drinking from a cup/glass/bottle & 145 & 0.2539 & 4.4276 & 11 & Closet / Walk-in closet / Spare closet & Opening a closet/cabinet & 121 & 0.0389 & 1.1810 \\
12 & bed & Lying on a bed & 335 & 0.3579 & 3.3828 & 12 & Closet / Walk-in closet / Spare closet & Tidying up a closet/cabinet & 81 & 0.0260 & 1.7496 \\
13 & laptop & Working/playing on a laptop & 436 & 0.2425 & 3.1496 & 13 & Bathroom & Washing their hands & 71 & 0.0182 & 2.3622 \\
14 & sandwich & Eating a sandwich & 570 & 0.3677 & 2.1848 & 14 & Bedroom & Lying on a bed & 275 & 0.0260 & 1.6405 \\
15 & light & Turning on a light & 280 & 0.4921 & 3.8159 & 15 & Entryway & Opening a door & 148 & 0.0424 & 0.9805 \\
\bottomrule
\end{tabular}
}
\end{table*}

For AAH, each row corresponds to an ordered action pair drawn from an event sequence. We build two complementary relation types conditioned on the current action: AA-prev, which measures how likely a previous action is given the current action, and AA-next, which measures how likely a next action is given the current action. Concretely, the benchmark uses $P(a_{i-1}\mid a_i)$ for predecessor priors and $P(a_{i+1}\mid a_i)$ for successor priors.
Tab.~\ref{tab:appendix_aa_examples} shows that these temporal priors are highly structured and semantically meaningful. Many high-probability pairs correspond to natural event scripts, such as closing being strongly associated with opening as its previous action, or putting back being strongly associated with taking as either a predecessor or successor depending on the query direction.
These examples clarify why AAH is particularly challenging: the hallucinated action is not only plausible in isolation, but also strongly licensed by the local temporal context surrounding the current action.

\begin{table*}[t]
\centering
\scriptsize
\caption{Representative directional action--action transition statistics used in AAH construction. AA-prev reports predecessor priors $P(a_{i-1}\mid a_i)$ and AA-next reports successor priors $P(a_{i+1}\mid a_i)$. Entries with conditional probability $=1.0$ are excluded.}
\label{tab:appendix_aa_examples}
\resizebox{\textwidth}{!}{%
\begin{tabular}{cllcc|cllcc}
\toprule
\multicolumn{5}{c|}{\textbf{AA-prev: Previous Action before the Current Action}} & \multicolumn{5}{c}{\textbf{AA-next: Next Action after the Current Action}} \\
\textbf{Rank} & \textbf{Current Action} & \textbf{Previous Action} & \textbf{Pair Count} & \textbf{$P(\text{prev}\mid\text{current})$} & \textbf{Rank} & \textbf{Current Action} & \textbf{Next Action} & \textbf{Pair Count} & \textbf{$P(\text{next}\mid\text{current})$} \\
\midrule
1 & Closing a book & Opening a book & 11 & 0.5000 & 1 & Opening a book & Closing a book & 14 & 0.9333 \\
2 & Closing a refrigerator & Opening a refrigerator & 11 & 0.5000 & 2 & Opening a refrigerator & Closing a refrigerator & 7 & 0.7778 \\
3 & Closing a closet/cabinet & Opening a closet/cabinet & 18 & 0.6429 & 3 & Opening a closet/cabinet & Closing a closet/cabinet & 16 & 0.6154 \\
4 & Closing a door & Opening a door & 13 & 0.5909 & 4 & Opening a door & Closing a door & 20 & 0.7143 \\
5 & Closing a box & Opening a box & 9 & 0.6000 & 5 & Opening a box & Closing a box & 8 & 0.8889 \\
6 & Holding some food & Opening a refrigerator & 7 & 0.3182 & 6 & Taking a broom from somewhere & Putting a broom somewhere & 6 & 0.8571 \\
7 & Putting a phone/camera somewhere & Taking a phone/camera from somewhere & 11 & 0.6111 & 7 & Holding a broom & Tidying something on the floor & 3 & 0.7500 \\
8 & Talking on a phone/camera & Taking a phone/camera from somewhere & 5 & 0.2778 & 8 & Taking a phone/camera from somewhere & Putting a phone/camera somewhere & 9 & 0.9000 \\
9 & Putting a towel/s somewhere & Taking a towel/s from somewhere & 11 & 0.5500 & 9 & Taking a towel/s from somewhere & Putting a towel/s somewhere & 9 & 0.9000 \\
10 & Putting a dish/es somewhere & Taking a dish/es from somewhere & 7 & 0.8750 & 10 & Taking a dish/es from somewhere & Putting a dish/es somewhere & 9 & 0.7500 \\
11 & Putting a cup/glass/bottle somewhere & Taking a cup/glass/bottle from somewhere & 27 & 0.4219 & 11 & Taking a cup/glass/bottle from somewhere & Putting a cup/glass/bottle somewhere & 22 & 0.5946 \\
12 & Drinking from a cup/glass/bottle & Taking a cup/glass/bottle from somewhere & 19 & 0.2969 & 12 & Taking a cup/glass/bottle from somewhere & Drinking from a cup/glass/bottle & 15 & 0.4054 \\
13 & Putting a cup/glass/bottle somewhere & Drinking from a cup/glass/bottle & 12 & 0.3871 & 13 & Drinking from a cup/glass/bottle & Putting a cup/glass/bottle somewhere & 9 & 0.6429 \\
14 & Putting some food somewhere & Taking food from somewhere & 21 & 0.5250 & 14 & Taking food from somewhere & Putting some food somewhere & 20 & 0.7143 \\
15 & Putting clothes somewhere & Taking some clothes from somewhere & 7 & 0.4118 & 15 & Holding some medicine & Taking/consuming some medicine & 4 & 0.5714 \\
\bottomrule
\end{tabular}%
}
\end{table*}

\subsection{Sequential Inference Construction Details}
Sequential inference hallucination (SIH) is designed around sports whose terminal outcome is highly stereotyped and therefore easily over-inferred from early visual cues. We select 11 sports in which a particular end-state almost always marks the completion of the action sequence: landing for jumping and gymnastics events, entering the water for diving, and release for throwing events. This design lets us truncate videos before the terminal sub-action while preserving strong trigger evidence for the eventual outcome. Tab.~\ref{tab:appendix_sih_sports} summarizes the selected sports and their corresponding end sub-actions used for truncation-based evaluation.

This construction yields particularly adversarial negatives. For instance, in high jump or pole vault, the athlete may already have completed the run-up, takeoff, and bar-clearing stages, making landing on the mat overwhelmingly likely from a commonsense perspective even when it is not visually present. Similarly, in diving, the board takeoff and mid-air rotation strongly cue water entry, while in throwing events, the wind-up and delivery pose strongly cue release. 
Therefore, SIH isolates a distinct failure mode beyond ordinary recognition error: the model does not merely misread frames, but extrapolates a plausible endpoint that has been deliberately withheld.

\begin{table}[!t]
\centering
\scriptsize
\caption{Sports used to construct SIH and the terminal sub-action removed from the negative clip.}
\label{tab:appendix_sih_sports}
\begin{tabular}{ll}
\toprule
\textbf{Sport} & \textbf{Annotated End Sub-action} \\
\midrule
Vault & Landing on the mat \\
Tumbling & Landing on the mat \\
Triple jump & Landing in the sand pit \\
Javelin throw & Releasing the javelin \\
Discus throw & Releasing the discus \\
Shot put & Releasing the shot \\
Hammer throw & Releasing the hammer \\
Pole vault & Landing on the mat \\
Long jump & Landing in the sand pit \\
High jump & Landing on the mat \\
Diving & Entering the water \\
\bottomrule
\end{tabular}
\end{table}

\subsection{Similarity Construction Details}

For semantic similarity hallucination (SSH), we organize the 500 atomic actions in HAA500 into a semantic hierarchy and then construct negative pairs from the same leaf cluster. Concretely, we prompt GPT to build a three-level hierarchy, with an optional fourth level when further subdivision is necessary, as shown in Fig.~\ref{fig:appendix_similarity_prompt}. The prompt explicitly instructs the model to favor action mechanics, motor patterns, object interaction, and functional similarity over coarse topical grouping. As a result, actions are not clustered under overly broad parents such as sports or playing instruments; instead, they are placed under increasingly specific superclasses that preserve fine-grained action resemblance. We then define semantic confounders as actions that fall within the same leaf cluster as the ground-truth action, thereby ensuring that the negative candidate is conceptually close yet still motion-distinct.

\begin{figure*}[t]
\centering
\caption{Prompt used to induce the semantic hierarchy over the 500 HAA500 action categories.}
\label{fig:appendix_similarity_prompt}
\fbox{%
\begin{minipage}{0.96\textwidth}
\footnotesize
\ttfamily
\raggedright
You are a classification assistant. I have a list of 500 action categories. Your task is to organize them into a multi-level hierarchical taxonomy (at least 3 levels, up to 4 levels when appropriate).\
\par
Hierarchy Requirements: Use fine-grained superclasses, avoiding overly broad categories. For example, do NOT group: ``playing guitar'', ``playing piano'', ``playing keyboard'' under a broad superclass like: ``playing musical instruments''. Instead, create progressively specific layers, based on action similarity, motor pattern similarity, object interaction similarity, functional similarity, physical movement similarity, and instrument structure similarity.\
\par
The hierarchy should follow this structure: Level 1: Broad domain; Level 2: instrument family or interaction type; Level 3: instrument-specific action group; Level 4 (if applicable): technique-level distinction. If a category logically belongs to multiple branches, list it under each relevant superclass. Avoid generic superclasses such as ``playing'', ``sports'', or ``activities''. Every category must be placed under the most specific logical superclass. Maintain consistent indentation to reflect hierarchy depth. Do not collapse similar but distinguishable actions. Be structurally consistent across all 500 categories. Favor action-mechanics similarity over purely semantic similarity.\
\par
Output Format:\
Level 1 Superclass:\
\quad Level 2 Superclass:\
\quad\quad Level 3 Superclass:\
\quad\quad\quad Subclass 1\
\quad\quad\quad Subclass 2\
\quad\quad\quad Subclass 3\
Repeat until all 500 categories are classified.\
\par
Example input prefix: id, name; 1,ALS IceBucket Challenge; 2,CPR; 3,abseiling;
\end{minipage}%
}
\end{figure*}

For visual similarity hallucination (VSH), we construct samples from multiple clips segmented from the same original HACS video. Given a target action text $a_{\text{true}}$, we first use Qwen3-VL-Reranker to score text--video relevance and manually verify the retrieved clips, ensuring that each positive clip contains $a_{\text{true}}$ while each negative clip does not. We then use the same reranker to compute video--video similarity between source-matched positive/negative clips, and retain pairs that are visually similar but semantically different in action content.
The resulting VSH pairs therefore share the same source context and similar visual appearance, while differing only in whether the target action is present.

\section{Evaluation Protocol and Additional Metrics}

\subsection{Generative Evaluation for Sequential Inference Hallucination}
For the open-ended SIH protocol, generations that explicitly contain lexical variants of the withheld negative action, such as landing, entering the water, or release, are directly counted as hallucinations. For the remaining generations, we use DeepSeek as a conservative entailment judge to determine whether the model output semantically implies that the omitted end sub-action has already happened. This second-stage check is necessary because models often hallucinate completion without copying the exact action string.

\paragraph{Prompt Templates}
The entailment prompt in Fig.~\ref{fig:appendix_sih_prompt} instructs the judge to decide whether a model-generated description entails that the negative action occurred. The prompt emphasizes three constraints: (1) this is a semantic entailment task rather than keyword matching; (2) preparatory or intermediate actions do not count as completion; and (3) when the evidence is ambiguous, the decision should default to ``UNCERTAIN'' rather than over-claiming hallucination. The final label set is ``ENTAILED'', ``NOT\_ENTAILED'', and ``UNCERTAIN'', which are mapped to a binary hallucination indicator by treating only ``ENTAILED'' as positive.

\begin{figure*}[t]
\centering
\caption{Prompt used to judge whether a generated SIH description entails the withheld terminal action.}
\label{fig:appendix_sih_prompt}
\fbox{%
\begin{minipage}{0.96\textwidth}
\footnotesize
\ttfamily
\raggedright
You are a strict semantic entailment judge for action occurrence.\\
\par
Your job is to determine whether MODEL\_OUTPUT entails that NEGATIVE\_ACTION has actually happened.\\
\par
Rules:\\
1. This is a semantic entailment task, not keyword matching.\\
2. Return ENTAILED only when MODEL\_OUTPUT explicitly states or strongly entails that NEGATIVE\_ACTION occurred.\\
3. Return NOT\_ENTAILED when MODEL\_OUTPUT describes only preparation, attempt, approach, partial execution, a related action, a broader action, or a description that does not support that NEGATIVE\_ACTION actually happened.\\
4. Return UNCERTAIN when the text is ambiguous or insufficient to determine whether NEGATIVE\_ACTION actually happened.\\
5. Do not assume completion from common knowledge or likely event sequences unless it is strongly entailed by the wording.\\
6. Judge only from the provided text.\\
7. Be conservative. If there is reasonable uncertainty between ENTAILED and NOT\_ENTAILED, choose UNCERTAIN.\\
\par
Label mapping:\\
- ENTAILED $\rightarrow$ is\_neg = true\\
- NOT\_ENTAILED $\rightarrow$ is\_neg = false\\
- UNCERTAIN $\rightarrow$ is\_neg = false\\
\par
Output valid JSON only in exactly this format:\\
\{"is\_neg": true, "label": "ENTAILED", "reason": "one-sentence explanation"\}\\
or\\
\{"is\_neg": false, "label": "NOT\_ENTAILED", "reason": "one-sentence explanation"\}\\
or\\
\{"is\_neg": false, "label": "UNCERTAIN", "reason": "one-sentence explanation"\}
\end{minipage}%
}
\end{figure*}

\begin{figure*}[t]
    \centering
    \includegraphics[width=\textwidth]{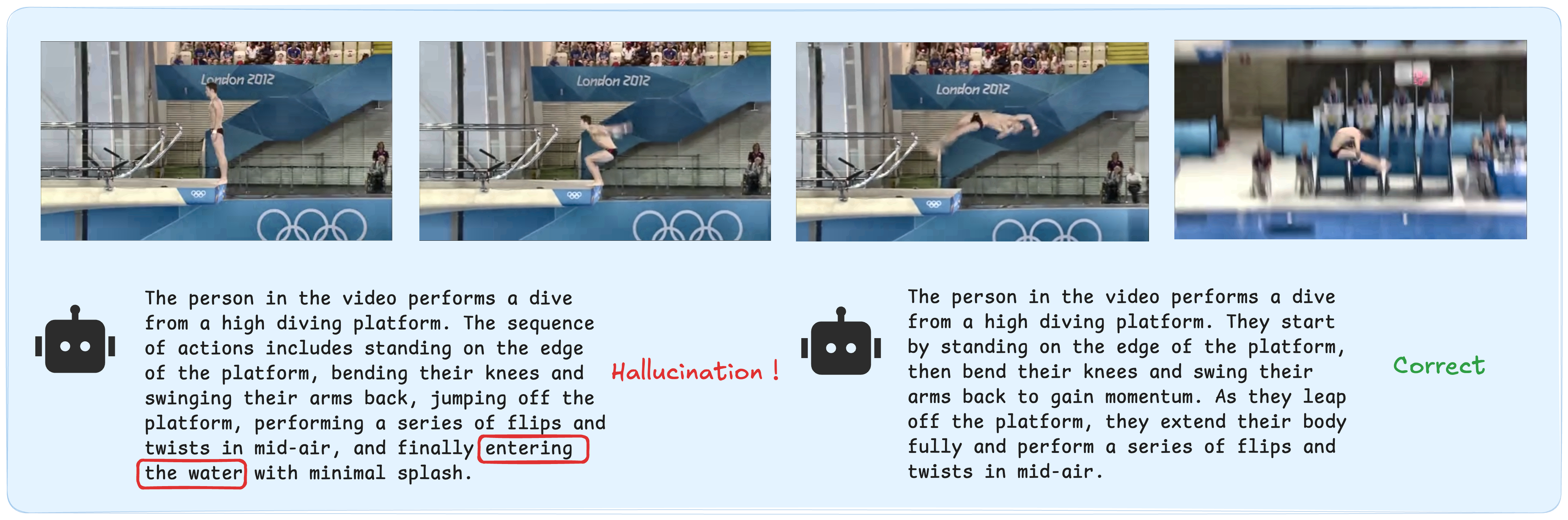}
    \caption{Illustration of explicit endpoint hallucination in the open-ended SIH setting. For a diving clip whose terminal action ``entering the water'' is removed, the left response is judged as hallucination because it explicitly entails the omitted endpoint, while the right response is judged as correct because it only describes the observed pre-outcome stages.}
    \label{fig:appendix_oq_example}
\end{figure*}

This design lets us detect both explicit and implicit endpoint hallucinations while keeping the decision criterion conservative.
Fig.~\ref{fig:appendix_oq_example} illustrates this distinction with a diving example. In the left case, the generated description explicitly states that the diver is ``entering the water,'' which directly entails the withheld terminal action and is therefore counted as a hallucination. In the right case, the description only mentions the takeoff and mid-air flipping stages, and neither explicitly states nor semantically implies water entry. Therefore it is judged as non-hallucinatory. 
These examples therefore illustrate the decision boundary of our protocol between true endpoint hallucination and grounded partial-event description.

\subsection{Additional Metrics}
The main paper reports $\mathrm{Acc}_{\mathrm{PS}}$, $\mathrm{Acc}_{\mathrm{NS}}$, $\mathrm{Cons}$, and $\mathrm{Q\mbox{-}PairAcc}$. To further probe model behavior in the binary-choice (BC) setting, the appendix introduces three complementary groups of diagnostics: four raw accuracies that describe performance under each question--sample configuration, a stricter pair-level metric, and sample-specific consistency metrics. Together, these metrics expose whether a model's apparent robustness comes from genuine discrimination or from asymmetric answer bias.

Let $\hat{y}_{i, q}^{+}$ and $\hat{y}_{i, q}^{-}$ denote the predictions for the positive and negative samples of the $i$-th adversarial pair under query polarity $q\in\{\text{pos},\text{neg}\}$, and let $y_{i, q}^{+}, y_{i, q}^{-}$ be the corresponding ground-truth labels used in the definitions below.

We first report four descriptive raw accuracies. $a_{\text{pos}}^{+}$ measures accuracy under the positive query on positive samples, where the correct answer is Yes. $a_{\text{pos}}^{-}$ measures accuracy under the positive query on negative samples, where the correct answer is No. $a_{\text{neg}}^{+}$ measures accuracy under the reversed query on positive samples, where the correct answer is No. Finally, $a_{\text{neg}}^{-}$ measures accuracy under the reversed query on negative samples, where the correct answer is Yes. These four numbers provide the most direct view of sample-type asymmetry and question-type asymmetry.

Beyond the $\text{Q-PairAcc}$ reported in the main paper, we introduce a stricter pair-level metric that requires all four binary judgments associated with the same adversarial pair to be correct:
\begin{equation}
\text{PairAcc} = \frac{1}{N}\sum_{i=1}^{N} \mathbb{I}\left[
\begin{aligned}
&\hat{y}_{i,\text{pos}}^{+}=y_{i,\text{pos}}^{+} \wedge \\
&\hat{y}_{i,\text{pos}}^{-}=y_{i,\text{pos}}^{-} \wedge \\
&\hat{y}_{i,\text{neg}}^{+}=y_{i,\text{neg}}^{+} \wedge \\
&\hat{y}_{i,\text{neg}}^{-}=y_{i,\text{neg}}^{-}
\end{aligned}
\right].
\end{equation}
Compared with $\mathrm{Q\mbox{-}PairAcc}$, $\mathrm{PairAcc}$ is a stricter final indicator of both anti-hallucination ability and logical coherence because it jointly requires correctness across positive and reversed question formulations as well as across positive and negative samples.

To further analyze logical consistency, we separate the consistency score used in the main paper into positive-sample and negative-sample components:
\begin{equation}
\text{ConsPS} = \frac{1}{N}\sum_{i=1}^{N} \mathbb{I}\left[\hat{y}_{i,\text{neg}}^{+} = 1 - \hat{y}_{i,\text{pos}}^{+}\right],
\end{equation}
\begin{equation}
\text{ConsNS} = \frac{1}{N}\sum_{i=1}^{N} \mathbb{I}\left[\hat{y}_{i,\text{neg}}^{-} = 1 - \hat{y}_{i,\text{pos}}^{-}\right].
\end{equation}
Here, $\mathrm{ConsPS}$ measures how often the model gives logically complementary answers when the same positive sample is queried with positive versus reversed wording, while $\mathrm{ConsNS}$ measures the analogous property on negative samples. These additional metrics reveal several interesting patterns that are invisible from headline accuracy alone: some models achieve strong raw accuracy on the easy quadrants but remain weak on full-pair correctness, while others exhibit markedly different consistency behavior on positive versus negative samples.

\section{Additional Experimental Details}

\subsection{Frame Sampling Ablation}
\begin{figure}[t]
    \centering
    \includegraphics[width=\linewidth]{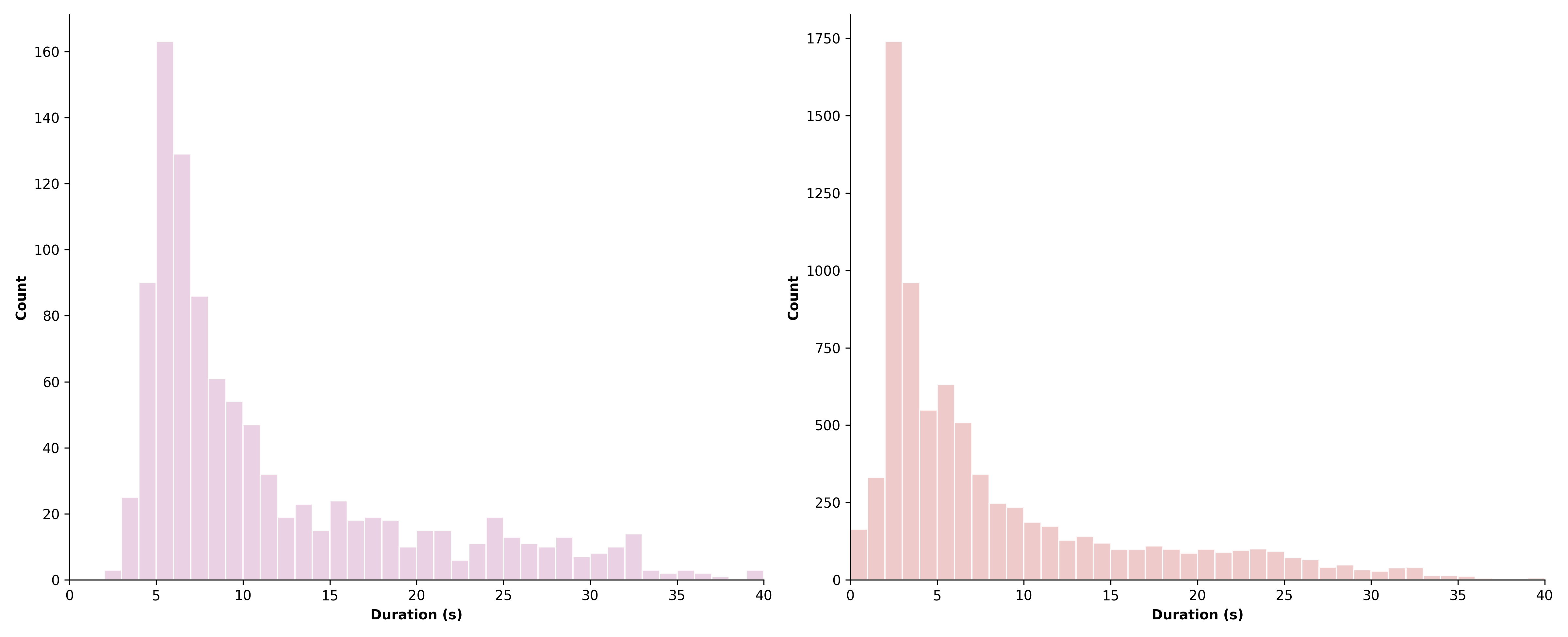}
    \caption{Clip-duration distributions of the OAH subset (left) and the full benchmark (right).} 
    \label{fig:appendix_duration_oah}
\end{figure}

We analyze the effect of the number of sampled frames using Qwen3-VL-8B-Instruct on OAH. This subset is especially informative because OAH videos are, on average, longer than the benchmark-wide mean and therefore impose a stricter temporal coverage requirement. Fig.~\ref{fig:appendix_duration_oah} compares the clip-duration distributions of the full benchmark and the OAH subset. OAH is visibly shifted toward longer clips, indicating that it demands stronger temporal coverage than the benchmark average. Consequently, if performance already saturates near 32 frames on OAH, the same sampling budget is a conservative and reasonable choice for the benchmark as a whole.

Tabs.~\ref{tab:appendix_frame_ablation} and \ref{tab:appendix_frame_mc} report the corresponding BC and MC results as the number of sampled frames increases on OAH.
The results reveal four clear trends. First, both MC accuracy in Tab.~\ref{tab:appendix_frame_mc} and BC accuracy in Tab.~\ref{tab:appendix_frame_ablation} improve in a pronounced stepwise manner as the frame count increases from 2 to 16, confirming that severe under-sampling harms temporal grounding. Second, performance is already near its peak around 32 frames: increasing the budget further to 64 or 128 brings little additional benefit, while substantially increasing token cost and computational burden. Third, the MC option distribution becomes more confident as the frame count increases: the Null rate drops sharply, but the hard-negative rate stays nearly unchanged, showing that extra frames reduce abstention without weakening the hallucination pull of the adversarial distractor. Fourth, more frames amplify yes bias. As the model sees more temporal evidence, $a_{\text{pos}}^{+}$ and $a_{\text{neg}}^{+}$ increase steadily, showing stronger recognition of positive samples under both question polarities; however, $a_{\text{pos}}^{-}$ and $a_{\text{neg}}^{-}$ decline markedly, indicating that the model becomes increasingly willing to assert that a plausible action has occurred even when it is absent. Finally, $\mathrm{ConsPS}$ and $\mathrm{ConsNS}$ plateau after roughly 4--8 frames, suggesting that logical self-consistency does not improve simply by extending the visual input sequence. In other words, longer temporal context helps recognition more than it helps reasoning.

\begin{table}[t]
\centering
\scriptsize
\caption{Qwen3-VL-8B-Instruct frame-sampling ablation on OAH. OAH contains longer videos on average, making it a stricter test of temporal coverage; performance saturates around 16--32 frames.}
\label{tab:appendix_frame_ablation}
\setlength{\tabcolsep}{3pt}
\resizebox{\columnwidth}{!}{%
\begin{tabular}{lcccccccccc}
\toprule
\textbf{Setting} & $a_{\text{pos}}^{+}$ & $a_{\text{pos}}^{-}$ & $a_{\text{neg}}^{+}$ & $a_{\text{neg}}^{-}$ & $\text{Acc}_{\text{PS}}$ & $\text{Acc}_{\text{NS}}$ & \textbf{ConsPS (\%)} & \textbf{ConsNS (\%)} & \textbf{Q-PairAcc} & \textbf{PairAcc (\%)} \\
\midrule
2 frames & 29.0 & \textbf{73.5} & 34.7 & \textbf{74.0} & 31.9 & \textbf{73.8} & 73.6 & \textbf{81.7} & 17.5 & 8.6 \\
4 frames & 49.4 & 62.7 & 60.4 & 60.7 & 54.9 & 61.7 & 70.0 & 74.7 & 28.1 & 14.7 \\
8 frames & 61.9 & 59.9 & 74.2 & 53.2 & 68.0 & 56.5 & 70.8 & 74.2 & \textbf{33.9} & 17.8 \\
16 frames & 63.8 & 57.0 & 80.4 & 46.2 & 72.1 & 51.6 & 72.2 & 73.2 & 33.2 & 18.1 \\
32 frames & 65.6 & 54.0 & 83.4 & 41.9 & 74.5 & 47.9 & 72.6 & 73.6 & 33.1 & \textbf{18.5} \\
64 frames & \textbf{67.8} & 51.8 & 84.9 & 38.7 & \textbf{76.4} & 45.2 & \textbf{73.7} & 72.0 & 31.8 & 18.0 \\
128 frames & 66.7 & 50.6 & \textbf{86.0} & 37.1 & \textbf{76.4} & 43.9 & \textbf{73.7} & 72.1 & 30.3 & 16.0 \\
\bottomrule
\end{tabular}
}
\end{table}

\begin{table}[t]
\centering
\scriptsize
\caption{Qwen3-VL-8B-Instruct MC frame-sampling ablation on OAH. We report the option distribution over the ground-truth answer (GT), hard negative (HN), random negative (RN), and null option.}
\label{tab:appendix_frame_mc}
\setlength{\tabcolsep}{4pt}
\resizebox{\columnwidth}{!}{%
\begin{tabular}{lcccc}
\toprule
\textbf{Setting} & \textbf{GT (\%)} & \textbf{HN (\%)} & \textbf{RN (\%)} & \textbf{Null (\%)} \\
\midrule
2 frames & 27.9 & \textbf{33.1} & \textbf{12.3} & \textbf{26.8} \\
4 frames & 40.3 & \textbf{33.1} & \textbf{11.6} & 15.1 \\
8 frames & 47.9 & 28.7 & 10.0 & 13.4 \\
16 frames & 51.9 & 27.0 & 10.0 & 11.1 \\
32 frames & \textbf{52.0} & 29.7 & 8.6 & 9.8 \\
64 frames & 51.0 & 30.2 & 8.0 & 10.9 \\
128 frames & 50.8 & 29.9 & 8.4 & 10.9 \\
\bottomrule
\end{tabular}%
}
\end{table}

\section{Additional Results}

\subsection{Binary-Choice Diagnostics}
We next report the BC diagnostic results under the metrics defined above. Tab.~\ref{tab:appendix_bc_diagnostic_ch} summarizes CH, and Tab.~\ref{tab:appendix_bc_diagnostic_sih_sh} summarizes SIH and SH.
The raw accuracies reveal which query--sample configurations are relatively easy, while $\mathrm{ConsPS}$, $\mathrm{ConsNS}$, and $\mathrm{PairAcc}$ reveal whether these local gains translate into globally coherent anti-hallucination behavior across adversarial settings.

\begin{table*}[t]
\centering
\scriptsize
\caption{Binary-choice (BC) diagnostics for co-occurrence hallucination (CH), reported with the bias-aware metrics defined above. Row groups correspond to object--action hallucination (OAH), scene--action hallucination (SAH), and action--action hallucination (AAH).}
\label{tab:appendix_bc_diagnostic_ch}
\resizebox{\textwidth}{!}{%
\begin{tabular}{lccccccc}
\toprule
\textbf{Model} & $a_{\text{pos}}^{+}$(\%) & $a_{\text{pos}}^{-}$(\%) & $a_{\text{neg}}^{+}$(\%) & $a_{\text{neg}}^{-}$(\%) & \textbf{ConsPS (\%)} & \textbf{ConsNS (\%)} & \textbf{PairAcc (\%)} \\
\midrule
\multicolumn{8}{c}{\textbf{OAH}} \\
\midrule
InternVL3-8B & 84.9 & 33.6 & 84.0 & 40.4 & 80.2 & 72.4 & 13.6 \\
LLaVA-Video-7B-Qwen2 & 71.0 & 49.2 & 98.2 & 9.6 & 72.4 & 55.1 & 3.2 \\
MiniCPM-V-4.5 & 98.1 & 9.7 & 41.9 & \textbf{70.7} & 42.5 & 35.5 & 3.0 \\
Molmo-2-8B & 71.8 & 70.1 & 98.4 & 10.4 & 72.7 & 37.6 & 6.5 \\
Perception-LM-8B & 78.6 & 60.1 & 82.7 & 36.9 & 77.3 & 65.9 & 17.8 \\
Qwen2.5-VL-7B-Instruct & 50.0 & \textbf{76.1} & \textbf{99.8} & 0.2 & 50.2 & 23.9 & 0.0 \\
Qwen2-VL-7B-Instruct & \textbf{99.5} & 1.3 & 74.8 & 50.6 & 75.3 & 50.4 & 0.9 \\
Qwen3-VL-8B-Instruct & 65.6 & 54.0 & 83.4 & 41.9 & 72.6 & 73.6 & \textbf{18.5} \\
Tarsier2-7B & 70.7 & 57.8 & 98.4 & 12.3 & 70.8 & 49.4 & 5.4 \\
VideoLlama3-7B & 93.2 & 25.0 & 88.7 & 31.8 & \textbf{87.5} & \textbf{74.7} & 11.9 \\
\midrule
\multicolumn{8}{c}{\textbf{SAH}} \\
\midrule
InternVL3-8B & 95.4 & 52.7 & 89.3 & 56.0 & \textbf{89.6} & \textbf{73.4} & 37.6 \\
LLaVA-Video-7B-Qwen2 & 82.1 & 72.0 & 96.9 & 20.7 & 85.2 & 48.8 & 14.4 \\
MiniCPM-V-4.5 & 99.2 & 28.9 & 58.3 & \textbf{70.2} & 58.3 & 47.6 & 14.0 \\
Molmo-2-8B & 85.4 & 84.4 & 97.2 & 31.1 & 87.6 & 46.7 & 23.6 \\
Perception-LM-8B & 84.9 & 83.9 & 88.9 & 60.3 & 88.1 & 68.1 & \textbf{46.2} \\
Qwen2.5-VL-7B-Instruct & 61.8 & \textbf{85.7} & \textbf{100.0} & 0.2 & 62.0 & 14.6 & 0.1 \\
Qwen2-VL-7B-Instruct & \textbf{100.0} & 11.6 & 80.5 & 66.0 & 80.5 & 45.2 & 7.8 \\
Qwen3-VL-8B-Instruct & 86.8 & 67.2 & 89.7 & 52.7 & 91.0 & 79.0 & 42.1 \\
Tarsier2-7B & 84.3 & 82.9 & 97.4 & 28.9 & 84.9 & 43.5 & 11.1 \\
VideoLlama3-7B & 93.6 & 49.8 & 90.3 & 53.0 & 84.3 & 72.8 & 25.8 \\
\midrule
\multicolumn{8}{c}{\textbf{AAH}} \\
\midrule
InternVL3-8B & 93.3 & 20.5 & 83.0 & 24.9 & 80.4 & 56.3 & 5.3 \\
LLaVA-Video-7B-Qwen2 & 77.0 & 44.0 & 93.6 & 8.1 & 78.2 & 38.6 & 2.3 \\
MiniCPM-V-4.5 & 97.8 & 6.8 & 48.7 & \textbf{57.0} & 49.1 & 44.1 & 1.7 \\
Molmo-2-8B & 80.3 & 51.9 & 91.7 & 10.6 & 76.0 & 58.0 & 6.4 \\
Perception-LM-8B & 88.9 & 49.6 & 83.7 & 33.5 & 80.1 & 61.8 & \textbf{15.8} \\
Qwen2-VL-7B-Instruct & \textbf{99.6} & 2.6 & 76.9 & 40.1 & 76.5 & 62.2 & 1.8 \\
Qwen2.5-VL-7B-Instruct & 49.1 & \textbf{70.0} & \textbf{94.1} & 0.4 & 47.4 & 30.9 & 0.2 \\
Qwen3-VL-8B-Instruct & 75.4 & 45.3 & 81.2 & 28.1 & 73.9 & 67.2 & 13.2 \\
Tarsier2-7B & 82.9 & 60.3 & 91.5 & 10.4 & 75.9 & 45.1 & 6.2 \\
VideoLlama3-7B & 97.1 & 24.7 & 84.5 & 28.4 & \textbf{84.4} & \textbf{71.6} & 11.9 \\
\bottomrule
\end{tabular}%
}
\end{table*}

\begin{table*}[t]
\centering
\scriptsize
\caption{Binary-choice (BC) diagnostics for sequential inference hallucination (SIH) and similarity hallucination (SH), reported with the bias-aware metrics defined above. Row groups correspond to SIH, semantic similarity hallucination (SSH), and visual similarity hallucination (VSH).}
\label{tab:appendix_bc_diagnostic_sih_sh}
\resizebox{\textwidth}{!}{%
\begin{tabular}{lccccccc}
\toprule
\textbf{Model} & $a_{\text{pos}}^{+}$(\%) & $a_{\text{pos}}^{-}$(\%) & $a_{\text{neg}}^{+}$(\%) & $a_{\text{neg}}^{-}$(\%) & \textbf{ConsPS (\%)} & \textbf{ConsNS (\%)} & \textbf{PairAcc (\%)} \\
\midrule
\multicolumn{8}{c}{\textbf{SIH}} \\
\midrule
InternVL3-8B & 99.9 & 0.8 & 99.6 & 10.4 & \textbf{99.5} & \textbf{89.4} & 0.3 \\
LLaVA-Video-7B-Qwen2 & 99.5 & 14.3 & 99.9 & 0.1 & 99.4 & 85.7 & 0.1 \\
MiniCPM-V-4.5 & 99.7 & 0.6 & 14.1 & \textbf{96.8} & 14.3 & 3.8 & 0.0 \\
Molmo-2-8B & 96.0 & 25.7 & \textbf{100.0} & 0.1 & 96.0 & 74.4 & 0.1 \\
Perception-LM-8B & 98.9 & 20.4 & 67.9 & 54.4 & 68.2 & 58.8 & 7.4 \\
Qwen2.5-VL-7B-Instruct & 92.6 & 28.0 & \textbf{100.0} & 0.0 & 92.6 & 72.1 & 0.0 \\
Qwen2-VL-7B-Instruct & \textbf{100.0} & 0.0 & 28.2 & 84.5 & 28.2 & 15.5 & 0.0 \\
Qwen3-VL-8B-Instruct & 99.7 & 4.0 & 91.8 & 24.9 & 92.1 & 77.9 & 2.7 \\
Tarsier2-7B & 98.4 & \textbf{47.7} & 98.4 & 12.4 & 97.8 & 59.9 & \textbf{9.3} \\
VideoLlama3-7B & 99.7 & 13.3 & 91.7 & 25.5 & 91.5 & 73.1 & 5.8 \\
\midrule
\multicolumn{8}{c}{\textbf{SSH}} \\
\midrule
InternVL3-8B & 94.5 & 30.9 & 95.3 & 23.3 & \textbf{91.9} & \textbf{76.2} & 13.9 \\
LLaVA-Video-7B-Qwen2 & 86.9 & 51.3 & 98.9 & 1.8 & 88.0 & 50.4 & 1.1 \\
MiniCPM-V-4.5 & 85.5 & 42.0 & 91.4 & 14.0 & 79.1 & 58.7 & 5.9 \\
Molmo-2-8B & 87.8 & 53.3 & 99.1 & 8.3 & 88.4 & 54.9 & 6.8 \\
Perception-LM-8B & 85.3 & 59.6 & 89.6 & 28.4 & 82.0 & 60.0 & 20.1 \\
Qwen2.5-VL-7B-Instruct & 53.5 & \textbf{80.4} & \textbf{100.0} & 0.0 & 53.7 & 19.9 & 0.0 \\
Qwen2-VL-7B-Instruct & \textbf{100.0} & 2.5 & 83.2 & \textbf{40.1} & 83.2 & 62.1 & 2.2 \\
Qwen3-VL-8B-Instruct & 86.4 & 55.8 & 90.2 & 33.6 & 89.1 & 73.2 & \textbf{25.9} \\
Tarsier2-7B & 93.2 & 50.1 & 95.3 & 10.8 & 90.7 & 57.0 & 7.6 \\
VideoLlama3-7B & 93.1 & 30.6 & 95.2 & 19.2 & 90.1 & 76.0 & 11.3 \\
\midrule
\multicolumn{8}{c}{\textbf{VSH}} \\
\midrule
InternVL3-8B & 95.8 & 25.6 & 95.2 & 36.1 & \textbf{94.5} & 73.0 & 15.5 \\
LLaVA-Video-7B-Qwen2 & 89.8 & 52.6 & 99.6 & 4.8 & 90.1 & 52.2 & 4.0 \\
MiniCPM-V-4.5 & 87.8 & 45.9 & 97.2 & 12.7 & 87.1 & 57.3 & 5.9 \\
Molmo-2-8B & 87.6 & 53.1 & 99.5 & 5.5 & 87.8 & 52.4 & 4.5 \\
Perception-LM-8B & 91.9 & 43.6 & 91.4 & 38.5 & 87.6 & 68.8 & 20.0 \\
Qwen2.5-VL-7B-Instruct & 73.1 & \textbf{67.6} & \textbf{100.0} & 0.2 & 73.3 & 33.6 & 0.1 \\
Qwen2-VL-7B-Instruct & \textbf{99.7} & 0.8 & 88.9 & \textbf{42.5} & 89.2 & 58.2 & 0.7 \\
Qwen3-VL-8B-Instruct & 88.6 & 50.9 & 94.5 & 34.3 & 90.0 & \textbf{74.1} & \textbf{23.6} \\
Tarsier2-7B & 88.1 & 59.1 & 98.5 & 13.6 & 88.7 & 52.9 & 11.1 \\
VideoLlama3-7B & 92.9 & 33.6 & 96.2 & 18.7 & 92.9 & 70.9 & 10.3 \\
\bottomrule
\end{tabular}%
}
\end{table*}

The BC diagnostics in Tab.~\ref{tab:appendix_bc_diagnostic_ch} and Tab.~\ref{tab:appendix_bc_diagnostic_sih_sh} reveal a consistent pattern across settings: many models remain strong on $a_{\text{pos}}^{+}$ and often also on $a_{\text{neg}}^{+}$, but are much weaker on $a_{\text{pos}}^{-}$, $a_{\text{neg}}^{-}$, and especially on $\mathrm{PairAcc}$. This indicates that current VideoLLMs are much better at affirming present motions than at rejecting plausible but absent ones. In other words, the dominant failure is motion falsification rather than motion recognition.
A second observation is that high consistency does not necessarily imply strong anti-hallucination ability. This is most evident in SIH, where several models still obtain near-zero $\mathrm{PairAcc}$ despite relatively high $\mathrm{ConsPS}$ and $\mathrm{ConsNS}$. This suggests that their errors are not merely caused by query framing, but by systematic over-inference from partial motion cues.


\begin{table*}[t]
\centering
\scriptsize
\caption{Multiple-choice (MC) option distributions for co-occurrence hallucination (CH) and similarity hallucination (SH). The top block reports OAH, SAH, and AAH, while the bottom block reports SSH and VSH, with VSH split into positive and negative subsets. GT, HN, RN, and Null denote the ground-truth answer, hard negative, random negative, and null option, respectively. For VSH-neg, Pos\_Action denotes selecting the action from the matched positive clip, and RN aggregates the two random-action options.}
\label{tab:appendix_mc}
\resizebox{\textwidth}{!}{%
\begin{tabular}{lcccc|cccc|cccc}
\toprule
\multirow{2}{*}{\textbf{Model}} & \multicolumn{4}{c|}{\textbf{OAH}} & \multicolumn{4}{c|}{\textbf{SAH}} & \multicolumn{4}{c}{\textbf{AAH}} \\
& \textbf{GT (\%)} & \textbf{HN (\%)} & \textbf{RN (\%)} & \textbf{Null (\%)} & \textbf{GT (\%)} & \textbf{HN (\%)} & \textbf{RN (\%)} & \textbf{Null (\%)} & \textbf{GT (\%)} & \textbf{HN (\%)} & \textbf{RN (\%)} & \textbf{Null (\%)} \\
\midrule
Molmo-2-8B & \textbf{75.0} & 13.1 & 9.4 & 2.4 & \textbf{78.2} & 13.6 & 4.9 & 3.3 & \textbf{72.1} & 20.4 & 7.5 & 0.1 \\
Tarsier2-7B & \textbf{73.7} & 13.9 & 9.5 & 2.9 & \textbf{82.4} & 9.9 & 5.1 & 2.6 & \textbf{80.8} & 13.5 & 5.6 & 0.2 \\
Perception-LM-8B & \textbf{73.6} & 9.3 & 8.0 & 9.2 & \textbf{83.5} & 5.0 & 3.4 & 8.1 & \textbf{80.8} & 13.6 & 3.7 & 1.9 \\
Qwen3-VL-8B-Instruct & \textbf{52.0} & 29.7 & 8.6 & 9.8 & \textbf{76.1} & 10.8 & 2.7 & 10.4 & \textbf{70.0} & 18.0 & 5.2 & 6.8 \\
Qwen2-VL-7B-Instruct & \textbf{69.4} & 14.9 & 14.5 & 1.2 & \textbf{71.0} & 18.5 & 7.3 & 3.2 & \textbf{72.3} & 19.2 & 7.9 & 0.7 \\
InternVL3-8B & \textbf{63.8} & 24.2 & 9.3 & 2.7 & \textbf{74.5} & 16.2 & 5.1 & 4.2 & \textbf{72.6} & 20.0 & 5.5 & 1.9 \\
VideoLlama3-7B & \textbf{62.1} & 19.9 & 15.2 & 2.8 & \textbf{53.8} & 17.6 & 8.5 & 20.1 & \textbf{63.1} & 24.6 & 9.0 & 3.3 \\
Qwen2.5-VL-7B-Instruct & \textbf{60.8} & 17.5 & 11.8 & 9.9 & \textbf{72.8} & 16.8 & 4.2 & 6.2 & \textbf{65.2} & 25.5 & 7.0 & 2.2 \\
MiniCPM-V-4.5 & \textbf{48.4} & 34.6 & 15.2 & 1.8 & \textbf{56.4} & 22.7 & 7.7 & 13.2 & \textbf{63.7} & 0.0 & 0.0 & 36.3 \\
LLaVA-Video-7B-Qwen2 & \textbf{47.6} & 39.6 & 11.3 & 1.4 & \textbf{77.5} & 15.7 & 5.3 & 1.6 & \textbf{64.0} & 25.2 & 10.0 & 0.8 \\
\bottomrule
\end{tabular}%
}

\resizebox{\textwidth}{!}{%
\begin{tabular}{lcccc|ccc|ccc}
\toprule
\multirow{2}{*}{\textbf{Model}} & \multicolumn{4}{c|}{\textbf{SSH}} & \multicolumn{3}{c|}{\textbf{VSH-pos}} & \multicolumn{3}{c}{\textbf{VSH-neg}} \\
& \textbf{GT (\%)} & \textbf{HN (\%)} & \textbf{RN (\%)} & \textbf{Null (\%)} & \textbf{GT (pos action, \%)} & \textbf{RN (\%)} & \textbf{Null (\%)} & \textbf{GT (None, \%)} & \textbf{Pos Action (\%)} & \textbf{RN (\%)} \\
\midrule
Perception-LM-8B & \textbf{84.8} & 5.3 & 0.0 & 0.6 & \textbf{99.3} & 0.1 & 0.7 & 11.0 & \textbf{87.9} & 1.1 \\
Qwen3-VL-8B-Instruct & \textbf{81.0} & 5.3 & 0.0 & 1.1 & \textbf{98.0} & 0.3 & 1.7 & 20.4 & \textbf{78.6} & 0.9 \\
Qwen2.5-VL-7B-Instruct & \textbf{79.3} & 7.2 & 0.0 & 1.1 & \textbf{98.9} & 0.3 & 0.8 & 14.3 & \textbf{84.2} & 1.5 \\
InternVL3-8B & \textbf{78.9} & 8.0 & 0.0 & 0.2 & \textbf{98.4} & 0.1 & 1.5 & 18.1 & \textbf{80.6} & 1.3 \\
Qwen2-VL-7B-Instruct & \textbf{78.7} & 7.0 & 0.0 & 0.0 & \textbf{99.2} & 0.6 & 0.2 & 4.7 & \textbf{93.0} & 2.3 \\
LLaVA-Video-7B-Qwen2 & \textbf{77.6} & 7.8 & 0.0 & 1.4 & \textbf{98.7} & 0.5 & 0.9 & 13.3 & \textbf{84.3} & 2.4 \\
Molmo-2-8B & \textbf{77.3} & 7.5 & 0.0 & 0.2 & \textbf{99.6} & 0.2 & 0.2 & 2.8 & \textbf{93.5} & 3.7 \\
MiniCPM-V-4.5 & \textbf{74.1} & 7.8 & 0.0 & 0.7 & \textbf{94.1} & 0.9 & 5.0 & 34.1 & \textbf{63.0} & 2.9 \\
VideoLlama3-7B & \textbf{72.5} & 8.9 & 0.0 & 2.5 & \textbf{95.3} & 0.5 & 4.2 & 25.6 & \textbf{72.9} & 1.5 \\
Tarsier2-7B & \textbf{87.5} & 4.8 & 0.0 & 0.0 & \textbf{99.2} & 0.4 & 0.4 & 9.6 & \textbf{88.4} & 2.0 \\
\bottomrule
\end{tabular}%
}
\end{table*}

\subsection{Multiple-Choice Option Distributions}
For CH and SH, the MC setting provides a complementary perspective on hallucination: instead of asking whether a single action occurred, we force the model to choose among the ground-truth option, a hard negative, a random negative, and a null answer. The resulting option distributions reveal whether the model fails by selecting the adversarial prior, by collapsing to null, or by drifting away from the correct option. VSH uses a different four-option design: one target action $a_{\text{true}}$, two random actions, and ``None of these.'' On positive clips, the correct answer is $a_{\text{true}}$; on negative clips, the correct answer is ``None of these.'' We therefore report VSH-pos and VSH-neg separately, and for VSH-neg we additionally isolate the proportion of errors that select the positive-clip action, which directly measures visually induced action hallucination.

The MC results in Tab.~\ref{tab:appendix_mc} are consistent with the BC findings. In CH, many errors come from direct attraction to the hard negative, especially in OAH and AAH, showing that prior-consistent distractors often dominate over actual motion evidence. In contrast, some models in AAH choose the null option much more frequently, suggesting a more conservative failure mode under sequential ambiguity.
The SH results show a different pattern. In SSH, errors are distributed across the hard negative, random negative, and null options, indicating difficulty in resolving fine-grained semantic neighbors even when the confounders are semantically constrained. In VSH, positive clips remain easy, but on negative clips models overwhelmingly choose the action associated with the matched positive clip rather than ``None of these.'' This shows that visual similarity exerts a strong hallucination-inducing effect and is the dominant source of error in VSH.

Overall, these additional results show that the three axes of MoHallBench expose different hallucination mechanisms. CH is dominated by hard-negative attraction, SIH by process completion, SSH by confusion among fine-grained semantic neighbors, and VSH by failures to reject visually similar but action-mismatched clips under strong appearance-level cues.

\end{document}